\pdfoutput=1
\documentclass[11pt]{article}

\usepackage[preprint]{acl}
\usepackage{times}
\usepackage{latexsym}

\usepackage[T1]{fontenc}
\usepackage[utf8]{inputenc}

\usepackage{microtype}
\usepackage{inconsolata}

\usepackage{hyperref}
\usepackage{url}

\usepackage{booktabs}
\usepackage{nicefrac}
\usepackage{xcolor}
\usepackage{graphicx}
\usepackage{wrapfig}
\usepackage{subcaption}
\usepackage{dirtytalk}
\usepackage{textcomp}
\usepackage{listings}
\usepackage{pythonhighlight}
\usepackage{amsmath}
\usepackage{amssymb}
\usepackage{amsthm}
\usepackage{amsfonts}
\usepackage{cleveref}
\usepackage{makecell}
\usepackage{multicol}
\usepackage{multirow}
\usepackage{bbding}
\usepackage{colortbl}
\usepackage{pifont}
\usepackage{mathrsfs}
\usepackage{bbm}
\usepackage{bm}
\usepackage{enumitem}
\usepackage{xspace}
\usepackage{dsfont}
\usepackage{soul}
\usepackage{tikz}
\usepackage{centernot}
\usepackage{tcolorbox}
\usepackage{tabularx}

\title{Improving Language Models with Intentional Analysis}

\author{Yuwei Yin \\ % \thanks{}
University of British Columbia \\
\texttt{yuweiyin@cs.ubc.ca} \\
\And
Giuseppe Carenini \\
University of British Columbia \\
\texttt{carenini@cs.ubc.ca} \\
}

\definecolor{ia_yellow}{rgb}{1.0, 1.0, 0.848}
\definecolor{ia_green}{rgb}{0.95, 1.0, 0.94}
\definecolor{ia_green_dark}{rgb}{0.88, 1.0, 0.88}

\begin{document}

\maketitle

\begin{abstract}
Intent, a critical cognitive notion and mental state, is ubiquitous in human communication and problem-solving. Accurately understanding the underlying intent behind questions is imperative to reasoning towards correct answers.
However, this significant concept has been largely disregarded in the rapid development of language models (LMs).
To unleash the potential of intent and instill it into LMs, this paper introduces Intentional Analysis (IA), which explicitly invokes intent-aware analysis and reasoning during the problem-solving process.
Comprehensive experiments across diverse benchmarks, model types, and configurations demonstrate the effectiveness, robustness, and generalizability of IA.
Notably, IA consistently improves task performance even on SOTA proprietary models like GPT-5 and Claude-Opus-4.6. Moreover, IA not only outperforms Chain-of-Thought (CoT) across various experimental settings, but it can also synergistically work with CoT reasoning.
Further qualitative analysis and case studies reveal that the benefits of IA stem from addressing several weaknesses in baseline methods, such as intent misunderstanding, hasty generalization, and mental laziness.
Case studies also provide insights into the mechanisms underlying IA and clarify how it differs from CoT in mitigating these weaknesses.
This study sheds light on a promising direction for the development of future LLMs with intentional analysis.\footnote{Source code: \url{https://github.com/YuweiYin/IA}}
\end{abstract}

%%%%%%%%%% # %%%%%%%%%% # SECTION # %%%%%%%%%% # %%%%%%%%%%
\section{Introduction}
\label{sec:introduction}

% 1. Intuition of IA
Intentional analysis, a meta-analysis of the given question regarding its intentionality~\citep{anscombe1956intention,sokolowski1984intentional,mele1989intention,mele1994intentional}, is critical to problem-solving because analyzing the question's intent leads to a thorough contextual understanding, a clear problem-solving target, and a purposeful planning guide.
On the contrary, solutions without intentional analysis may fail to capture the genuine intents and implications, mislead subsequent reasoning, and thus result in incorrect answers.
Based on this intuition, we propose improving language models (LMs) by instilling Intentional Analysis (IA) into these systems for general problem-solving.

% 2. Background of LM
Despite the revolutionary breakthroughs of LMs~\citep{zhao2023llm_survey,min2023llm_survey,minaee2024llm_survey} in the field of Natural Language Processing (NLP), their further advancement faces obstacles when dealing with more and more complicated problems.
Existing methods for boosting the model performance mainly focus on LM pre-training scaling~\citep{kaplan2020scaling,openai2026gpt5}, data cleansing~\citep{soldaini2024dolma,penedo2024fineweb}, post-training techniques~\citep{ouyang2022rlhf,rafailov2023dpo,guo2025deepseek_r1_nature}, and test-time scaling~\citep{zhang2025tts_survey,muennighoff2025tts_s1}, with step-by-step Chain-of-Thought training and prompting~\citep{wei2022cot,kojima2022cot_think_step_by_step,jaech2024openai_o1} dominating the development of LM reasoning strategies.

% 3. Our hypotheses
However, intent, an ubiquitous cognitive notion in all kinds of problems, questions, or queries, is largely ignored by the current LM research.
As understanding the underlying intents is arguably always useful to problem-solving, we hypothesize that \ding{202} IA would contribute to performance gains for various types of LMs on a wide range of tasks, \ding{203} IA should perform on par with CoT, as both are general problem-solving methods, and \ding{204} IA could work synergistically with CoT and enhance the performance of CoT, since the basic idea of IA is fundamentally orthogonal to that of CoT: IA highlights intent understanding and intentional analysis, while CoT emphasizes step-by-step reasoning.

% 4.1. IA prompting - implementation
In this work, we empirically validate our three hypotheses with extensive experimental results and in-depth analysis, by exploring both inference-time IA prompting and IA fine-tuning strategies.
Specifically, we instruct the model to conduct explicit intentional analysis via a simple prompt: ``\textit{Let's analyze the intent of the question and then answer.}''.
Owing to the autoregressive text-completion nature~\citep{vaswani2017transformer,openai2019gpt2} and instruction-following capabilities~\citep{ouyang2022rlhf,wei2022sft} of recent LMs~\citep{grattafiori2024llama3,openai2026gpt5},
IA prompting (IA-PT) guides the model to first understand the question's intents and then leads to an intentional reasoning phase, where the reasoning process is based on a deeper intentional comprehension.
This behavior makes the whole solution more purposeful and pertinent, and thus more likely to give a correct final answer, as we can observe in the case studies provided in \S~\ref{sec:analysis}.

% 4.2. IA prompting - experiments
Experiments on multiple benchmarks covering diverse task types, target domains, and input-output structures~\citep{hendrycks2021mmlu,joshi2017triviaqa,press2023bamboogle,suzgun2023bbh,austin2021mbpp} demonstrate the effectiveness of IA prompting, compared with several representative baseline methods of LM reasoning, planning, and memory recalling~\citep{kojima2022cot_think_step_by_step,wang2023plan,yasunaga2024analogical}.
Further experiments on varied generation settings, including IA prompt variants, LM generation temperatures, and random seeds, verify the robustness of IA against specific implementation details and varying experimental configurations.
In addition, we apply IA to various LMs of different training stages (including Base, Instruct, SFT, DPO, and RLVR stages)~\citep{grattafiori2024llama3,lambert2025tulu3}, model sizes, and model types (both open-weight and proprietary LMs)~\citep{yang2024qwen2_5,jiang2023mistral,almazrouei2023falcon}.
Tellingly, IA consistently outperforms CoT and improves the task performance by a large margin on SOTA proprietary LLMs, including Gemini-3-Flash~\citep{team2023gemini,team2025gemini2_5}, GPT-5.2~\citep{openai2026gpt5}, and Claude-Opus-4.6~\citep{anthropic2026claude_opus_4_6}
Notably, the results also show that IA not only outperforms CoT in various experimental settings, but it can also synergistically work with CoT to further improve performance.
Extensive results validate our three hypotheses and corroborate the contributions of IA, shedding light on a promising direction for the development of future LLMs.

% 5. IA fine-tuning - implementation and experiments
Encouraged by the promising results with inference-time IA prompting,
we also explore an IA fine-tuning (IA-FT) method to enhance the intentional analysis ability of the original LMs.
As none of the existing LMs have previously trained towards this objective, we propose specifically fine-tuning LMs on problem-solving solutions with intentional analysis, so that the fine-tuned model can better perform the IA process and solve the problems more accurately.
To this end, we construct instructional fine-tuning data by asking LMs to solve the problems in the training set, where their solutions should contain intentional analysis, 
and then conduct supervised fine-tuning with language modeling objectives.
The experimental results show that IA fine-tuning further boosts the model performance on different tasks using two different LMs.

% 6. Further analysis
Further analysis and case studies reveal critical issues in the direct generation baseline without IA (e.g., intent misunderstanding, hasty generalization, and mental laziness), provide insights into the working mechanism of IA, and investigate how IA differs from CoT in mitigating these issues.
Overall, IA is a novel, simple, and effective methodology that incorporates a critical cognitive concept (i.e., intent) into problem-solving for improving and developing next-generation language models.

%%%%%%%%%% # %%%%%%%%%% # SECTION # %%%%%%%%%% # %%%%%%%%%%
\section{Background}
\label{sec:background}

\paragraph{Intentional Analysis.}
Although intent plays a key role in human communication~\citep{adams1986intention,grosz1986attention} and resides ubiquitously in almost all sorts of problems and dialogues, existing intent-related research~\citep{louvan2020intent_survey,weld2022intent_survey} in the field of NLP primarily focuses on understanding the intent behind user queries by recognizing~\citep{cohen2025small,tang2025kapa}, classifying~\citep{park2025dynamic,sali2025navigating}, and clustering~\citep{lin2025spill,hong2025dial} predefined intents or discovering new ones~\citep{lin2020discovering,zhang2022new}.
In this work, we initiate the incorporation of intentional analysis, as free-form text generation, for problem-solving with LMs.

\begin{figure*}[t!]
  \centering
  \includegraphics[width=0.98\textwidth]{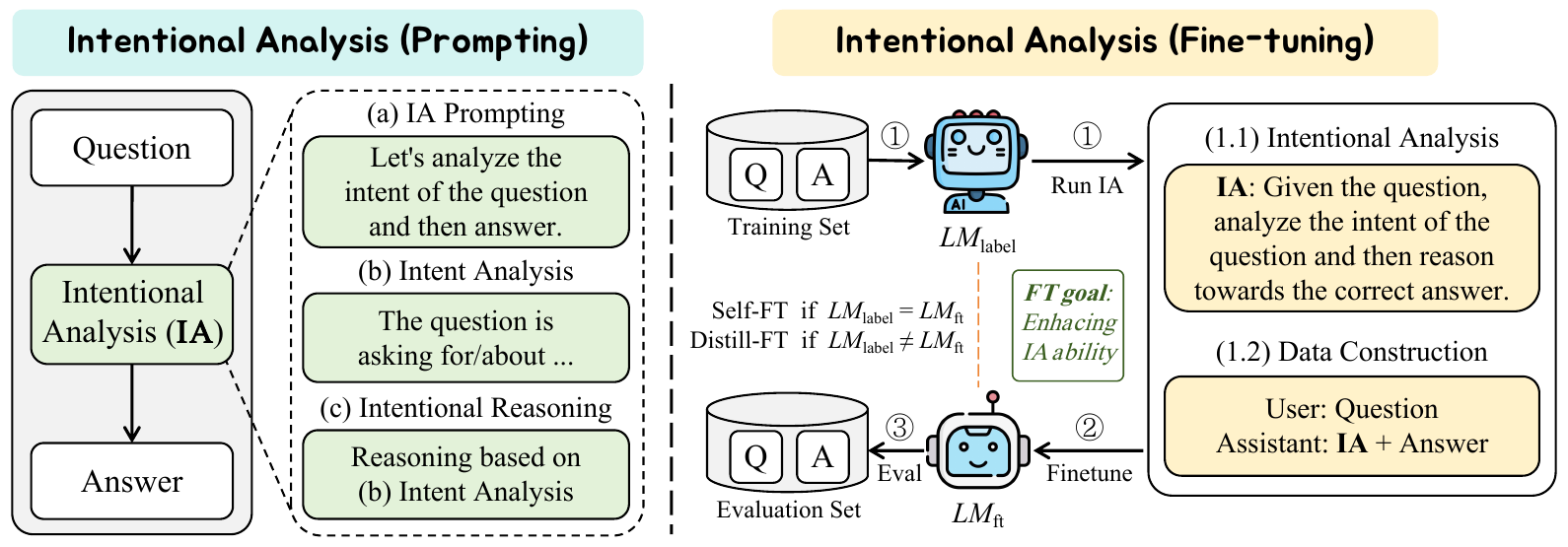}
  \vspace{-3pt}
  \caption{\textbf{Overview of Intentional Analysis (IA).} \textbf{\textit{IA Prompting}} (Left): The IA method requires the language model to first understand the question intent and then perform intentional reasoning based on the intent analysis. \\ \textbf{\textit{IA Fine-tuning}} (Right): After performing the IA method on the training set by the labeler model $LM_{\text{label}}$, we construct conversational style data and then fine-tune the target model $LM_{\text{ft}}$ in a self-improving or distillation way, where fine-tuning with IA solutions aims to enhance the intentional analysis ability of the original LM.}
  \label{fig:ia_overview}
  \vspace{-10pt}
\end{figure*}

\paragraph{Language Models.}
Large language models (LLMs)~\citep{zhao2023llm_survey,min2023llm_survey,minaee2024llm_survey} have been a transformative technique in NLP owing to their excellent text generation and conversation abilities~\citep{openai2026gpt5}.
Through pre-training on an enormous volume of data curated from the Internet~\citep{soldaini2024dolma,penedo2024fineweb,weber2024redpajama} and various post-training techniques like instructional supervised fine-tuning~\citep{ouyang2022rlhf,wei2022sft,zhang2025instruction_tuning_survey} and preference alignment~\citep{bai2022rlaif,rafailov2023dpo,guo2025deepseek_r1_nature}, LLMs excel at diverse tasks covering general world knowledge~\citep{joshi2017triviaqa,hendrycks2021mmlu}, complex logical reasoning~\citep{srivastava2023big_bench}, programming~\citep{austin2021mbpp}, and many more.
In this work, we propose to improve language models by eliciting and enhancing their ability to explicitly analyze intentions, which is underexplored in the current development of language models.

%%%%%%%%%% # %%%%%%%%%% # SECTION # %%%%%%%%%% # %%%%%%%%%%
\section{Intentional Analysis}
\label{sec:method}

In this section, we introduce our Intentional Analysis (IA) method, which can be implemented through LM prompting and fine-tuning approaches.

\subsection{Task Formulation}
\label{sec:method_task_formulation}

Formally, let $\mathcal{D} = \{\mathcal{Q}, \mathcal{A}\}$, where $\mathcal{Q}_i$ denotes the $i$-th question and $\mathcal{A}_i$ is the corresponding correct answer.
Depending on the task type, the ``question'' set $\mathcal{Q}$ and ``answer'' set $\mathcal{A}$ can mean differently.
For multiple-choice question answering (MCQA) tasks, each $\mathcal{Q}_i$ contains the question and options, and $\mathcal{A}_i$ is the correct option.
For open-ended question answering (OpenQA) tasks, each $\mathcal{Q}_i$ is the question, sometimes with additional context information, and $\mathcal{A}_i$ is a set of correct answers.
For coding tasks, each $\mathcal{Q}_i$ is a programming problem, usually with specific functional and formatting requirements, and $\mathcal{A}_i$ is a set of unit tests.

Regardless of the task types, the data type of questions and answers is free-form text, which will be tokenized by the model $LM$ as lists of integer token ids.
The output of $LM$ is also token ids and will be detokenized into free-form text for assessing the correctness (Specific details in \S~\ref{sec:exp_setup_evaluation}).

\subsection{Prompting LMs to Perform IA}
\label{sec:method_ia_prompting}

As illustrated in Figure~\ref{fig:ia_overview} (Left), we can require the model to perform Intentional Analysis (IA) by prompting it with a simple instruction: ``\textit{Let's analyze the intent of the question and then answer.}''.

With the instruction-following abilities~\citep{ouyang2022rlhf,wei2022sft} of recent language models~\citep{grattafiori2024llama3,openai2026gpt5}, IA prompting (IA-PT) can guide the model to better understand the question's intent through various means, such as paraphrasing the question in its own words, decomposing it, and extracting key components, among others.
Furthermore, owing to the autoregressive text-completion nature~\citep{vaswani2017transformer,openai2019gpt2} of language models, IA prompting leads to an intentional reasoning phase, where the reasoning process is based on a deeper understanding of the question's intent, making the analysis more purposeful and pertinent, and thus is hypothesized to enhance task performance.

\subsection{Fine-tuning LMs to Better Perform IA}
\label{sec:method_ia_finetune}

While IA prompting can intuitively improve the performance of language models on various tasks, the current LMs have not been specifically trained to analyze the intent of questions. Hence, we further propose to fine-tune LMs on IA-style problem-solving data to enhance their IA ability.

Let $\mathcal{D} = \{D_t, D_e\}$ be a dataset with a training set $D_t$ and an evaluation set $D_e$, where $D_t \cap D_e = \varnothing$. For datasets where $D_t=\varnothing$, we will only run LM evaluation and skip fine-tuning.
The training objective is classic next-token prediction~\citep{openai2019gpt2}, i.e., minimizing the average cross-entropy loss~\citep{shannon1951entropy,jurafsky2025slp} over the distributions of the model's predicted tokens and that of the expected tokens.
Here, the key is to construct fine-tuning data.

As shown in Figure~\ref{fig:ia_overview} (Right), to conduct IA fine-tuning, we first run the labeler model $LM_{\text{label}}$ on the training set $D_t$.
Specifically, given the question and ground-truth answer, $LM_{\text{label}}$ is asked to analyze the intent of the question and then reason towards the correct answer.
Letting each output solution (including IA prompting, intent analysis, and intentional reasoning) be $\mathcal{S}_i$, we construct the fine-tuning data in a conversational style: the user prompt is the question $\mathcal{Q}_i$, and the assistant prompt is the concatenation of $\mathcal{S}_i$ and $\mathcal{A}_i$.
During the evaluation on set $D_e$, the fine-tuned model $LM_{\text{ft}}$ will adopt IA prompting for problem-solving by default.

If $LM_{\text{label}}$ is the original version of $LM_{\text{ft}}$, we call it self-improving fine-tuning (Self-FT); if $LM_{\text{label}}$ is a different model (usually a larger and more powerful one), it is distillation fine-tuning (Distill-FT).
The model training and inference details in the experiments are elaborated in \S~\ref{sec:exp_setup}.

%%%%%%%%%% # %%%%%%%%%% # SECTION # %%%%%%%%%% # %%%%%%%%%%
\section{Experimental Setup}
\label{sec:exp_setup}

In this section, we introduce the experimental configurations that implement our IA method (\S~\ref{sec:method}) and set up the experiments.
More experiment details are elaborated in Appendix~\ref{app:experiment_detail}.

\subsection{Dataset and Evaluation Settings}
\label{sec:exp_setup_evaluation}

\paragraph{Datasets.}
As shown in Table~\ref{tab:datasets}, we consider a wide range of high-quality datasets, covering various task types and domains, input-output structures, and LM abilities to evaluate.
MMLU~\citep{hendrycks2021mmlu} comprehensively evaluates the language understanding ability of LMs on 50+ subtasks.
TriviaQA~\citep{joshi2017triviaqa} (wiki subset) contains Wikipedia-based knowledge-intensive questions.
Bamboogle~\citep{press2023bamboogle} is specifically designed for evaluating the compositionality ability of LMs.
BBH~\citep{srivastava2023big_bench,suzgun2023bbh} includes diverse reasoning-intensive questions of 20+ subtasks.
MBPP~\citep{austin2021mbpp} consists of Python programming problems and corresponding unit tests.

\paragraph{Evaluation.}
During evaluation, we extract the final answer~\citep{kojima2022cot_think_step_by_step} from the output text of the LM to compute the correctness of each instance.
For the coding task MBPP, we run the generated code on the unit tests and compute the Pass@1 score~\citep{chen2021humaneval}.
For other tasks, we apply exact matching (EM) on the normalized model prediction and the correct answer.
The final reported Accuracy score is the average of the correctness ratio per instance.

\begin{table}[t!]
    \centering
    \scalebox{0.7}{
    \begin{tabular}{cccc}
    \toprule
    \midrule
    \textbf{Dataset} & \textbf{Category} & \textbf{Structure} & \textbf{\textbf{Metric}} \\
    \midrule
    MMLU & General & MCQA & EM \\
    % TriviaQA (wiki) 
    TriviaQA & Knowledge & Single-hop OpenQA & EM \\
    Bamboogle & Compositionality & Multi-hop OpenQA & EM \\
    BBH & Reasoning & MCQA & EM \\
    MBPP & Coding & Programming & Pass@1 \\
    \midrule
    \bottomrule
    \end{tabular}
    }
    \vspace{-4pt}
    \caption{Datasets overview.}
    \label{tab:datasets}
    \vspace{-10pt}
\end{table}

\subsection{Baseline Settings}
\label{sec:exp_setup_baseline}

\subsubsection{Prompting Methods}
\label{sec:exp_setup_baseline_prompting}

Let the prompting instruction be $\mathcal{P}$. During evaluation, the user prompt of LM is the question $\mathcal{Q}_i$ and the assistant prompt is $\mathcal{P}$, so that the LM (i.e., assistant) can follow the instruction $\mathcal{P}$ and perform the required analysis.
For our \textbf{IA} prompting (\textbf{IA-PT}) method, $\mathcal{P}_{\text{IA}}=$ ``\textit{Let's analyze the intent of the question and then answer.}''
In \S~\ref{sec:results_prompting_effectiveness}, we also demonstrate that IA remains effective irrespective of its specific prompt formulation, as long as the model faithfully performs the IA process.
The basic baseline is \textbf{DA} (Direct Answer), where we do not apply any prompting. In other words, $\mathcal{P}_{\text{DA}}$ is an empty string. Outperforming the DA baseline shows the effectiveness of a prompting method.

We compare IA with several representative prompting baselines featured by reasoning, planning, and knowledge-retrieval.
\textbf{CoT} (Chain-of-Thought) prompting~\citep{kojima2022cot_think_step_by_step} is a prevalent reasoning prompting method and widely adopted in model training and inference, which requires LMs to perform step-by-step thinking.
\textbf{PS} (Plan-and-Solve)~\citep{wang2023plan} is a representative planning prompting method that requires models to plan before solving the problem.
\textbf{AR} (Analogical Reasoning)~\citep{yasunaga2024analogical} requires the model to recall its memory (i.e., parametric knowledge) before problem-solving.

\begin{table*}[t!]
    \centering
    \scalebox{0.75}{
    \begin{tabular}{cl|ccccc|l}
    \toprule
    \midrule
    & \multirow{2}{*}{\textbf{Method}} & General & Knowledge & Compositionality & Reasoning & Coding & \multirow{2}{*}{\textbf{Avg.}$_{(\Delta)}$} \\
    \cmidrule(lr){3-3} \cmidrule(lr){4-4} \cmidrule(lr){5-5} \cmidrule(lr){6-6} \cmidrule(lr){7-7}
    & & \textbf{MMLU} & \textbf{TriviaQA} & \textbf{Bamboogle} & \textbf{BBH} & \textbf{MBPP} & \\
    \midrule
    \rowcolor{lightgray}\ding{192} & DA (Direct Answer) & 67.71 & 76.93 & 44.80 & 56.02 & 38.72 & 56.84 \\
    \ding{193} & AR (Analogical Reasoning) & 67.83 & 77.99 & \textbf{47.20} & 56.92 & 46.32 & 59.25\textcolor{teal}{$_{(+2.41)}$} \\
    \ding{194} & PS (Plan-and-Solve) & \underline{69.12} & \underline{79.18} & 45.60 & 59.26 & \underline{47.35} & 60.10\textcolor{teal}{$_{(+3.26)}$} \\
    \ding{195} & CoT (Chain-of-Thought) & 68.74 & 78.96 & 46.40 & \underline{61.02} & 47.03 & \underline{60.43}\textcolor{teal}{$_{(+3.59)}$} \\
    \rowcolor{ia_yellow}\ding{196} & \textbf{IA} (Intentional Analysis) & \textbf{69.26} & \textbf{79.42} & \underline{46.93} & \textbf{61.17} & \textbf{48.93} & \textbf{61.13}\textcolor{teal}{$_{(+4.29)}$} \\
    \midrule
    \bottomrule
    \end{tabular}
    }
    \vspace{-4pt}
    \caption{\textbf{Effectiveness of Intentional Analysis}. The performance (Accuracy\%) of various baselines and our IA method on representative benchmarks requiring diverse capabilities.}
    \label{tab:exp_ia_effectiveness}
    \vspace{-5pt}
\end{table*}

\subsubsection{Fine-tuning Methods}
\label{sec:exp_setup_baseline_finetune}

As mentioned in \S~\ref{sec:method_ia_finetune}, we further boost their IA ability by fine-tuning LMs on IA-style solutions.
For our IA Fine-tuning (\textbf{IA-FT}) method, the user prompt is the question $\mathcal{Q}_i$, and the assistant prompt is the concatenation of IA solution $\mathcal{S}^{\text{IA}}_i$ and correct answer $\mathcal{A}_i$, where $\mathcal{S}^{\text{IA}}_i$ consists of IA prompting, intent analysis, and intentional reasoning.

\paragraph{Raw-FT.}
The Raw Fine-tuning (Raw-FT) baseline constructs training data with only the question $\mathcal{Q}_i$ (used in the user prompt) and correct answer $\mathcal{A}_i$ (used in the assistant prompt). I.e., the intermediate solution $\mathcal{S}^{\text{Raw}}_i$ is an empty string.
This baseline is to ablate the basic effect of LM fine-tuning.

\paragraph{DA-FT.}
The data construction process of the DA Fine-tuning (DA-FT) baseline runs DA prompting on the training set, obtains the analysis and reasoning data as $\mathcal{S}^{\text{DA}}_i$, and builds the assistant prompt as the concatenation of DA solution $\mathcal{S}^{\text{DA}}_i$ and the correct answer $\mathcal{A}_i$.
This baseline is to compare the effect of different reasoning solutions ($\mathcal{S}^{\text{DA}}_i$ or $\mathcal{S}^{\text{IA}}_i$) for LM fine-tuning.
Additionally, this baseline ablates the effect of the volume of fine-tuning data since the number of training tokens for DA-FT is close to (or more than) that for our IA-FT method.

%%%%%%%%%% # %%%%%%%%%% # SECTION # %%%%%%%%%% # %%%%%%%%%%
\section{Results}
\label{sec:results}

In this section, we report the experimental results, which demonstrate the effectiveness, robustness, and generalizability of Intentional Analysis.

For reproducibility, we strictly manage the software package versions, random seeds, LM generation temperatures, and other hyperparameters.
We set the generation temperature to $0$ by default and disable token sampling for deterministic generation.
Furthermore, we conducted all the generation experiments at least twice and obtained reproducible results.
Our source code is available at \href{https://github.com/YuweiYin/IA}{\textcolor{blue}{GitHub}}.

\subsection{Prompting LMs to Perform IA}
\label{sec:results_prompting}

\subsubsection{Effectiveness of Intentional Analysis}
\label{sec:results_prompting_effectiveness}

Table~\ref{tab:exp_ia_effectiveness} presents the performance of different methods on diverse benchmarks using the LLaMA3-8B-Instruct model~\citep{grattafiori2024llama3}.
The results demonstrate the effectiveness of our \ding{196} IA method, which consistently and significantly outperforms \ding{192} DA (Direct Answer), with t-test p-value less than $0.05$.
Compared to other representative baselines on parametric knowledge recalling \ding{193} AR (Analogical Reasoning)~\citep{yasunaga2024analogical}, LM planning \ding{194} PS (Plan-and-Solve)~\citep{wang2023plan}, and LM reasoning \ding{195} CoT (Chain-of-Thought)~\citep{kojima2022cot_think_step_by_step}, IA surpasses all the baselines on average.
By eliciting intentional analysis for problem-solving, our IA method performs the best in terms of Knowledge, Compositionality, and Coding categories, and scores the second best on General and Reasoning task types.

\begin{table*}[t!]
    \centering
    \scalebox{0.75}{
    \begin{tabular}{cllllllll}
    \toprule
    \midrule
    & \multirow{2}{*}{\textbf{Method}} & \multicolumn{5}{c}{\textbf{LLaMA3-8B} (of different training stages)} & \textbf{Mistral0.3-7B} & \textbf{Falcon3-7B} \\
    \cmidrule(lr){3-7} \cmidrule(lr){8-8} \cmidrule(lr){9-9}
    & & \textbf{Base} & \textbf{Instruct} & \textbf{Tulu3-SFT} & \textbf{Tulu3-DPO} & \textbf{Tulu3-RLVR} & \textbf{Instruct} & \textbf{Instruct} \\
    \midrule
    \ding{192} & \cellcolor{lightgray}DA & \underline{41.88} & 56.84 & \underline{51.29} & 56.53 & 55.63 & 46.81 & 52.34 \\
    \midrule
    \ding{193} & CoT & 38.05\textcolor{red}{$_{(-3.83)}$} & \underline{60.43}\textcolor{teal}{$_{(+3.59)}$} & 50.52\textcolor{red}{$_{(-0.77)}$} & \underline{59.58}\textcolor{teal}{$_{(+3.05)}$} & \underline{58.55}\textcolor{teal}{$_{(+2.92)}$} & \underline{48.96}\textcolor{teal}{$_{(+2.15)}$} & \underline{58.90}\textcolor{teal}{$_{(+6.56)}$} \\
    \midrule
    \ding{194} & \cellcolor{ia_yellow}\textbf{IA} & \textbf{43.66}\textcolor{teal}{$_{(+1.78)}$} & \textbf{61.13}\textcolor{teal}{$_{(+4.29)}$} & \textbf{54.46}\textcolor{teal}{$_{(+3.17)}$} & \textbf{61.00}\textcolor{teal}{$_{(+4.47)}$} & \textbf{58.89}\textcolor{teal}{$_{(+3.26)}$} & \textbf{50.08}\textcolor{teal}{$_{(+3.27)}$} & \textbf{59.36}\textcolor{teal}{$_{(+7.02)}$} \\
    \midrule
    \bottomrule
    \end{tabular}
    }
    \vspace{-4pt}
    \caption{\textbf{Generalizability of Intentional Analysis (with Open Models)}. Intentional Analysis consistently improves various open-weight language models of different training stages, model sizes, and model series.}
    \label{tab:exp_ia_generalizability_open}
    \vspace{-5pt}
\end{table*}

\begin{table}[t!]
    \centering
    \scalebox{0.75}{
    \begin{tabular}{cllll}
    \toprule
    \midrule
    & \textbf{Method} & \textbf{Gemini-3} & \textbf{GPT-5} & \textbf{Claude-Opus-4} \\
    \midrule
    \ding{192} & \cellcolor{lightgray}DA & 85.50 & 63.67 & 84.07 \\
    \midrule
    \ding{193} & CoT & \underline{85.85}\textcolor{teal}{$_{(+0.35)}$} & \underline{69.86}\textcolor{teal}{$_{(+6.19)}$} & \underline{85.08}\textcolor{teal}{$_{(+1.01)}$} \\
    \midrule
    \ding{194} & \cellcolor{ia_yellow}\textbf{IA} & \textbf{86.91}\textcolor{teal}{$_{(+1.41)}$} & \textbf{71.71}\textcolor{teal}{$_{(+8.04)}$} & \textbf{86.23}\textcolor{teal}{$_{(+2.16)}$} \\
    \midrule
    \bottomrule
    \end{tabular}
    }
    \vspace{-4pt}
    \caption{\textbf{Generalizability of Intentional Analysis}. Intentional Analysis consistently improves SOTA proprietary LLMs of different model series.}
    \label{tab:exp_ia_generalizability_closed}
    \vspace{-10pt}
\end{table}

\subsubsection{Robustness of Intentional Analysis}
\label{sec:results_prompting_robustness}

To demonstrate that IA is effective irrespective of specific prompt design, we consider the default IA prompt and five paraphrased variants (See Appendix~\ref{app:prompt_var}).
The consistent results in Table~\ref{tab:exp_ia_prompt_var} verify that IA works effectively regardless of its particular prompt formulation, so long as the model performs intentional analysis as instructed.
Furthermore, to explore the robustness of our IA method under varied LM generation settings, we investigate the performance perturbations when tuning the key hyperparameters controlling generation randomness: the temperature and random seeds.
The results in Table~\ref{tab:exp_gen_temp} and Table~\ref{tab:exp_random_seed} (See Appendix~\ref{app:results_prompting_robustness}) show that IA brings consistent gains under varying generation configurations, verifying its robustness.

\subsubsection{Generalizability of Intentional Analysis}
\label{sec:results_prompting_generalizability}

We further test IA on diverse language models to investigate the generalizability of IA.
Table~\ref{tab:exp_ia_generalizability_open} shows the performances on multiple benchmarks using different open-weight LMs:
% \ding{202} 
LLaMA3-8B models~\citep{grattafiori2024llama3} of different training stages, including Base, Instruct, SFT~\citep{lambert2025tulu3}, DPO~\citep{rafailov2023dpo}, and RLVR~\citep{guo2025deepseek_r1_nature}, along with open-weight LMs of different sizes~\citep{jiang2023mistral,almazrouei2023falcon}.
All the open-weight LMs are evaluated on MMLU, TriviaQA, Bamboogle, BBH, and MBPP datasets.
Furthermore, Table~\ref{tab:exp_ia_generalizability_closed} presents the performance of state-of-the-art (SOTA) proprietary LLMs: Google Gemini-3-Flash~\citep{team2023gemini,team2025gemini2_5}, OpenAI GPT-5.2~\citep{openai2026gpt5}, and Anthropic Claude-Opus-4.6~\citep{anthropic2026claude_opus_4_6}.
The proprietary models are tested on Bamboogle and GPQA~\citep{rein2023gpqa}, which are small-scale (thus cost-efficient) and still challenging benchmarks.

Compared with CoT, the second-best method in Table~\ref{tab:exp_ia_effectiveness}, 
our IA method shows dominance on all the open-weight and proprietary models, as shown in Table~\ref{tab:exp_ia_generalizability_open} and Table~\ref{tab:exp_ia_generalizability_closed}.
Moreover, CoT sometimes even underperforms the DA baseline (LLaMA3-8B-Base and LLaMA3-8B-Tulu3-SFT), while our IA method surpasses the DA baseline for all the LMs, demonstrating a solid benefit of conducting intentional analysis before reasoning and answering.
Notably, IA consistently outperforms CoT on SOTA models and improves the performance by a large margin.
Overall, extensive experiments on a wide range of model stages, sizes, and series solidify the generalizability of our IA method.

\subsubsection{Why is IA different from CoT?}
\label{sec:why_ia_vs_cot}

\paragraph{CoT idea vs. IA idea.}
In general, CoT requires LLMs to think step by step, utilizing the autoregressive nature of decoder-only language models, so that the final answer is generated based on the preceding reasoning chain. Despite being effective, CoT does not specify how LLMs should think or reason. In contrast, our IA method specifically emphasizes intentional understanding, which requires LLMs to first understand the intent and then answer based on the intentional analysis. Intuitively, a deep and clear understanding of intents should lead to a more accurate and focused solution. Thus, the basic idea of IA is distinct from that of CoT.

\paragraph{CoT performance vs. IA performance.}
Comprehensive experiments in \S~\ref{sec:results} demonstrate that IA outperforms CoT in diverse settings, i.e., different benchmarks, model types, model sizes, and model series. Moreover, IA exhibits consistent gains over the DA baseline, showing that generation with IA is better than without, while CoT sometimes underperforms the DA baseline (as in Table~\ref{tab:exp_ia_generalizability_open}).
Potentially, many recent LLMs have been exposed to CoT-style training data, yet their Intentional Analysis capabilities lag. Our findings highlight a promising direction for the further development of language models.

\begin{table}[t!]
    \centering
    \scalebox{0.75}{
    \begin{tabular}{ccccc}
    \toprule
    \midrule
    & \textbf{Model} & \textbf{CoT} & \cellcolor{ia_yellow}\textbf{IA} & \textbf{IA+CoT} \\
    \midrule
    \ding{192} & LLaMA3-8B-Base & 38.05 & \textbf{43.66} & \underline{40.28} \\
    \ding{193} & LLaMA3-8B-Instruct & 60.43 & \underline{61.13} & \textbf{62.25} \\
    \ding{194} & Tulu3-SFT & 50.52 & \textbf{54.46} & \underline{53.71} \\
    \ding{195} & Tulu3-DPO & 59.58 & \underline{61.00} & \textbf{61.01} \\
    \ding{196} & Tulu3-RLVR & 58.55 & \underline{58.89} & \textbf{59.46} \\
    \ding{197} & Mistral0.3-7B & 48.96 & \underline{50.08} & \textbf{50.42} \\
    \ding{198} & Falcon3-7B & 58.90 & \underline{59.36} & \textbf{60.29} \\
    \midrule
    \ding{199} & \textit{Average over models} & 53.57 & \textbf{55.51} & \underline{55.35} \\
    \midrule
    \bottomrule
    \end{tabular}
    }
    \vspace{-4pt}
    \caption{\textbf{Synergy between IA and CoT.} IA can not only beat CoT but also synergistically work with CoT.}
    \label{tab:exp_ia_plus_cot}
    \vspace{-10pt}
\end{table}

\paragraph{Synergy between IA and CoT.}
As indicated above, the ideas of IA and CoT are \textit{\textbf{orthogonal}}, i.e., IA highlights intent understanding and intentional analysis for problem-solving, while CoT emphasizes step-by-step reasoning.
Thus, to further substantiate the contribution of IA, we show that IA can not only beat CoT but also synergistically work with CoT.
Table~\ref{tab:exp_ia_plus_cot} reports performance comparisons among CoT, IA, and IA+CoT on the same five benchmarks as Table~\ref{tab:exp_ia_effectiveness}, where IA+CoT asks the model to \textit{analyze the intent of the question and then answer the question with step-by-step reasoning}.
As observed, the combined method IA+CoT consistently surpasses CoT, showing that IA can bring additional gains over CoT. For some models, IA alone is even better than the combined method, meaning that applying CoT sometimes leads to performance degradation, which might be due to its overthinking tendency.
The findings indicate the blind spots that CoT (step-by-step reasoning) exists, while IA (intentional analysis) could remedy.

\subsection{Fine-tuning LMs to Better Perform IA}
\label{sec:results_finetune}

Encouraged by the positive results of our IA method at inference time, and also inspired by the huge success of reasoning models~\citep{jaech2024openai_o1} that train LMs with step-by-step thinking data, we propose fine-tuning LMs to enhance their IA ability and inherently incorporate IA into them.

In the experiments, we fine-tune two different models $LM_{\text{ft}}$ (LLaMA3-8B-Instruct and Qwen2.5-7B-Instruct) on MMLU and TriviaQA, two datasets with large-scale training sets.
As mentioned in \S~\ref{sec:method_ia_finetune}, we consider self-improving and distillation fine-tuning.
The \textbf{IA-FT-self} and \textbf{DA-FT-self} methods apply self-improving fine-tuning, where the labeler model $LM_{\text{label}}$ is the same as the model to be fine-tuned $LM_{\text{ft}}$.
For distillation fine-tuning settings (i.e., \textbf{IA-FT-distill} and \textbf{DA-FT-distill}), Qwen3-32B~\citep{yang2025qwen3} is adopted as the labeler model $LM_{\text{label}}$.
Details of fine-tuning configurations are in Appendix~\ref{app:finetuning_details}.

\paragraph{Self-Improvement.}
As shown in Table~\ref{tab:exp_ia_finetune}, IA fine-tuning by itself (\ding{196} IA-FT-self) further boosts the performance of IA prompting (\ding{193} IA-PT), verifying the positive contribution of adapting the original LM to problem-solving with intentional analysis.
Compared to the DA-PT baseline, the relatively poor performance of \ding{194} Raw-FT and \ding{195} DA-FT-self methods implies that conducting fine-tuning itself or increasing the number of training tokens does not necessarily bring improvement, which substantiates the positive contributions of our IA fine-tuning method.
Among the fine-tuning methods, \ding{194} Raw-FT performs the worst, probably due to the lack of intermediate reasoning towards the final answer in its training process.
Besides, \ding{195} DA-FT-self performs close to \ding{192} DA-PT, possibly because training on its naturally generated reasoning traces does not change its behavior.

\vspace{-3pt}
\paragraph{Distillation.}
When adopting a larger and more powerful LM as the training set labeler, \ding{198} IA-FT-distill achieves additional, though minimal, performance gains compared to \ding{196} IA-FT-self.
Meanwhile, \ding{197} DA-FT-distill sometimes underperforms \ding{195} DA-FT-self, which means imitating the reasoning trace from the teacher model does not necessarily benefit the student model.
This might be because the style of problem analysis and reasoning from the teacher model can be hard to grasped by the student model, even though the general problem-solving ability of teacher is stronger.

\begin{table}[t!]
    \centering
    \scalebox{0.62}{
    \begin{tabular}{clcc|cc|c}
    \toprule
    \midrule
    \multicolumn{2}{c}{\multirow{2}{*}{\textbf{Method}}} & \multicolumn{2}{c}{LLaMA3-8B-Instruct} & \multicolumn{2}{|c|}{Qwen2.5-7B-Instruct} & \multirow{2}{*}{\textbf{Avg.}} \\
    \cmidrule(lr){3-4} \cmidrule(lr){5-6}
    & & \textbf{MMLU} & \textbf{TriviaQA} & \textbf{MMLU} & \textbf{TriviaQA} & \\
    \midrule
    \ding{192} & \cellcolor{lightgray}DA-PT & 67.71 & 76.93 & 70.83 & 67.31 & 70.70 \\
    \cmidrule{2-7}
    \ding{193} & \cellcolor{ia_yellow}\textbf{IA}-PT & 69.26 & 79.42 & 73.18 & 67.45 & 72.33 \\
    \midrule
    \midrule 
    \ding{194} & Raw-FT & 62.94 & 70.18 & 70.36 & 60.25 & 65.93 \\
    \cmidrule{2-7}
    \ding{195} & DA-FT-self & 66.31 & 77.14 & 71.39 & 67.58 & 70.61 \\
    \cmidrule{2-7}
    \ding{196} & \cellcolor{ia_green_dark}\textbf{IA}-FT-self & \underline{69.74} & \textbf{80.12} & \underline{73.94} & \underline{68.16} & \underline{72.99} \\
    \cmidrule{2-7}
    \ding{197} & DA-FT-distill & 66.52 & 75.33 & 71.72 & 67.02 & 70.15 \\
    \cmidrule{2-7}
    \ding{198} & \cellcolor{ia_green_dark}\textbf{IA}-FT-distill & \textbf{70.05} & \underline{79.72} & \textbf{74.33} & \textbf{68.33} & \textbf{73.11} \\
    \midrule
    \bottomrule
    \end{tabular}
    }
    \vspace{-4pt}
    \caption{\textbf{Fine-tuning LMs with Intentional Analysis}. The performance (Accuracy\%) of different prompting (PT, \ding{192}-\ding{193}) and fine-tuning (FT, \ding{194}-\ding{198}) methods.}
    \label{tab:exp_ia_finetune}
    \vspace{-10pt}
\end{table}

%%%%%%%%%% # %%%%%%%%%% # SECTION # %%%%%%%%%% # %%%%%%%%%%
\section{Analysis}
\label{sec:analysis}

Extensive results in \S~\ref{sec:results} have demonstrated the effectiveness, robustness, and generalizability of IA.
Here, we further analyze the model outputs of IA and the baseline methods to elaborate on IA's novel contributions, its distinction from baselines, and insights into its success and future development.

\subsection{How does IA work differently from DA?}
\label{sec:how_ia_vs_da}

To investigate the working mechanism of IA, We conduct a multi-step qualitative analysis of the outputs of our IA approach and the DA baseline (i.e., direct generation without IA).

\paragraph{IA Behavior.}
Observing the outputs using IA, we find that most of the content begins with restating, rewriting, or paraphrasing the question, with most of the outputs starting with ``The question is asking''.
As such, ruminating on the question's intent in the model's own words seems to enable the model to better grasp the ultimate goal and then more accurately reason towards the final answer.

\paragraph{Baseline Behavior.}
In contrast, the baseline methods either hastily jump into conclusions, along with post-hoc analysis as justification, or directly start solving the problem step by step. However, the reasoning path may sometimes lead to the wrong target without an accurate intentional understanding, no matter how strong the model's inherent reasoning capability is.

\paragraph{Case Studies.}
Furthermore, case studies in Appendix~\ref{app:case_study} reveal problems in the baseline method, such as \textit{intent misunderstanding}, \textit{hasty generalization}, \textit{mental laziness}, and \textit{faulty reasoning}.
Instead, our IA method appears to mitigate the issues with its intentional solution.
Specifically, we observe the following differences between IA and DA:
\begin{itemize}
\setlength\itemsep{0em}
    \item On general tasks from MMLU (Table~\ref{tab:case_study_mmlu}), the DA baseline often rushes to conclusions and then tries to present a post-hoc explanation, while our IA method begins with restating the question and detailing the key points, followed by point-by-point analysis.
    \item On knowledge-intensive tasks from TriviaQA and Bamboogle (See Table~\ref{tab:case_study_trivia_qa} and Table~\ref{tab:case_study_bamboogle} for examples), the intentional analysis stage helps the model better retrieve its own parametric knowledge, instead of recalling the incorrect memory or giving up trying (``\textit{mental laziness}''). With IA, LM would not simply respond ``I don't know'' even when it actually knows the answer.
    \item On reasoning-intensive tasks from BBH (Table~\ref{tab:case_study_bbh}) and programming tasks from MBPP (Table~\ref{tab:case_study_mbpp}), intentional analysis at the beginning helps with the subsequent reasoning and coding by specifying the correct direction and proposing feasible solutions.
\end{itemize}

\subsection{How does IA work differently from CoT?}
\label{sec:how_ia_vs_cot}

To mitigate the issues of the DA baseline, as revealed in \S~\ref{sec:how_ia_vs_da}, our IA method works differently from CoT. The prominent findings are as follows:

(1) CoT can greatly mitigate the \textit{mental laziness} issue of the DA baseline, but it still sometimes suffers from \textit{intent misunderstanding} and \textit{hasty generalization}. In contrast, our IA method excels at intent understanding and largely avoids \textit{hasty generalization} and \textit{mental laziness}. Still, both CoT and IA may suffer from the \textit{faulty reasoning} problem when the model's inherent reasoning ability is insufficient for complicated questions.

(2) When encountering hard questions, as an unstructured generic reasoning approach, CoT sometimes falls into an endless loop, repeating the same reasoning chains until hitting the max token limit, which is ineffective and inefficient. In comparison, our IA method requires models to first think through the intent of questions, so the generated intent analysis can serve as a guideline and lead to a purposeful solution, avoiding incessant self-doubting or self-correction.

\subsection{Further Discussions}
\label{sec:analysis_further}

We compare IA with additional related work \textbf{MI} (Mistral-Interact)~\citep{qian2024in3_mi}, a fine-tuned Mistral-7B~\citep{jiang2023mistral} model on the Intention-in-Interaction (IN3) benchmark, which inspects users' implicit intentions through explicit queries. Table~\ref{tab:exp_ia_vs_mi} in Appendix~\ref{app:exp_ia_vs_mi} shows that IA outperforms MI on all the tested benchmarks.
In addition, the fact that MI+IA consistently surpasses MI aligns with our findings on the synergy between IA and CoT (\S~\ref{sec:why_ia_vs_cot}).

Furthermore, we provide an efficiency study on the extra inference-time cost of IA, investigate which tasks IA would be more beneficial for, and present detailed examples of the model outputs in Appendix~\ref{app:efficiency_study}, Appendix~\ref{app:ia_subtask}, and Appendix~\ref{app:case_study}, respectively.
These findings provide useful clues regarding the working mechanism and benefits of IA, as well as suitable tasks and scenarios for applying IA, and suggest research directions for future work on improving LMs with intentional analysis.

%%%%%%%%%% # %%%%%%%%%% # SECTION # %%%%%%%%%% # %%%%%%%%%%
\section{Conclusion}
\label{sec:conclusion}

In this work, we introduce IA, an intuitive, effective, and general problem-solving method that improves the performance of language models by eliciting intentional analysis at inference time and further enhancing the model's IA ability through fine-tuning.
Comprehensive experiments on various benchmarks, language models, and experimental configurations, corroborate the effectiveness, robustness, and generalizability of our IA method.

In light of these positive results, IA opens a new path for developing next-generation LMs by building key cognitive concepts (e.g., intent) into the problem-solving process.
Specifically, the observations that (1) IA outperforms CoT and (2) IA can boost CoT performance draw attention to building intentional language models (using IA intentional solutions) in addition to large reasoning models (using CoT reasoning chains).

As a next step, we plan to dive into the development of intentional LLMs and investigate the IA paradigm in a broader context and application, such as multimodal and multicultural intentional understanding and analysis.

% \clearpage

%%%%%%%%%% # %%%%%%%%%% # SECTION # %%%%%%%%%% # %%%%%%%%%%
\section*{Limitations}
\label{sec:limitations}

Due to resource limits, we could not exhaustively evaluate all the available language models on all datasets, but we have experimented with our IA method on diverse benchmarks of different domains and structures using representative models of various series, model sizes, and training stages.

Regarding the experimental details, we could not enumerate all the possible hyperparameters for LM generation and fine-tuning. Thus, the reported performance could be further optimized through more delicate prompt optimization (for prompting) and fine-grained hyperparameter searching (for fine-tuning).
Still, our experiments on varied generation settings, including IA prompt variants, LM generation temperatures, and random seeds, verify the robustness of IA against specific implementation details and varying experimental configurations.

In terms of research scope, this work adopts text-only LMs instead of multimodal ones, but humans' multimodal understanding of the world is closely related to the comprehension and expression of intents.
Additionally, this work deals with English-only text, while different cultures (represented by different languages) may understand intent distinctly.
Thus, we would encourage future work to extend intentional analysis to multimodal and multicultural AI systems.

%%%%%%%%%% # %%%%%%%%%% # SECTION # %%%%%%%%%% # %%%%%%%%%%
\bigskip
\section*{Acknowledgments}
% Use unnumbered third level headings for the acknowledgments. All acknowledgments, including those to funding agencies, go at the end of the paper.

Nous remercions le Conseil de recherches en sciences naturelles et en g\'{e}nie du Canada (CRSNG) de son soutien. \\
We acknowledge the support of the Natural Sciences and Engineering Research Council of Canada (NSERC).
This research was supported in part by the computational resources and services provided by Advanced Research Computing at the University of British Columbia and the Digital Research Alliance of Canada (alliancecan.ca).
We would also like to thank Chuyuan Li (\texttt{chuyuan.li@ubc.ca}) for constructive feedback on the paper.

\newpage

\bibliography{ia}

\clearpage
\appendix

\section{Experiment Details}
\label{app:experiment_detail}

\subsection{Dataset Details}
\label{app:dataset_details}

All the datasets used in this work are loaded from Hugging Face datasets.
Table~\ref{tab:app_dataset_details} lists the source, split, and size of each dataset. The ``train'' split is used for model training, and other splits are used for model evaluation. Please note that the URLs may be subject to change by the dataset providers.

\begin{table}[ht]
    \centering
    \scalebox{0.7}{
    \begin{tabular}{llcc}
    \toprule
    \midrule
    \textbf{Dataset} & \textbf{URL} & \textbf{split} & \textbf{size} \\
    \midrule
    \multirow{2}{*}{MMLU~\citep{hendrycks2021mmlu}} & \multirow{2}{*}{\href{https://huggingface.co/datasets/cais/mmlu}{Link}} & train & 99,842 \\
    & & test & 14,042 \\
    \cmidrule(lr){2-4}
    \multirow{2}{*}{TriviaQA~\citep{joshi2017triviaqa}} & \multirow{2}{*}{\href{https://huggingface.co/datasets/mandarjoshi/trivia_qa}{Link}} & train & 61,888 \\
    & & valid & 7,993 \\
    \cmidrule(lr){2-4}
    Bamboogle~\citep{press2023bamboogle} & \href{https://huggingface.co/datasets/RUC-NLPIR/FlashRAG_datasets}{Link} & test & 125 \\
    BBH~\citep{suzgun2023bbh} & \href{https://huggingface.co/datasets/lukaemon/bbh}{Link} & test & 5,511 \\
    MBPP~\citep{austin2021mbpp} & \href{https://huggingface.co/datasets/google-research-datasets/mbpp}{Link} & test & 500 \\
    GPQA~\citep{rein2023gpqa} & \href{https://huggingface.co/datasets/Idavidrein/gpqa}{Link} & test & 198 \\
    \midrule
    \bottomrule
    \end{tabular}
    }
    \caption{The dataset details.}
    \label{tab:app_dataset_details}
\end{table}

\subsection{Model Details}
\label{app:model_details}

\paragraph{Language Models.}
In this work, we mainly adopt the instruction-following autoregressive language model LLaMA3-8B-Instruct~\citep{grattafiori2024llama3} to evaluate the effectiveness of IA (\S~\ref{sec:results_prompting_effectiveness}) and to analyze the benefits of IA (\S~\ref{sec:analysis}).
Aside from it, we also conduct extensive experiments on diverse LMs~\citep{jiang2023mistral,almazrouei2023falcon,lambert2025tulu3} of various series, model sizes, and training stages to verify the generalizability of IA (\S~\ref{sec:results_prompting_generalizability}).
All the open-weight models are loaded from HuggingFace~\citep{wolf2020transformers}, and the proprietary models are used via API calls with default configurations, e.g., the generation temperature for GPT models~\citep{openai2026gpt5} is 1.

All the open-weight language models used in this work are loaded from Hugging Face models.
Table~\ref{tab:app_model_source} lists the source of each model and tokenizer provided by Hugging Face Transformers~\citep{wolf2020transformers}.
Please note that the URLs may be subject to change by the model providers.

\begin{table}[ht]
    \centering
    \scalebox{0.7}{
    \begin{tabular}{lccl}
    \toprule
    \midrule
    \multicolumn{1}{c}{\textbf{LM Series}} & \textbf{Size} & \textbf{Type} & \textbf{URL} \\
    \midrule
    \multirow{5}{*}{LLaMA3~\citep{grattafiori2024llama3}} & 8B & Base & \href{https://huggingface.co/meta-llama/Llama-3.1-8B}{Link} \\
    & 8B & Instruct & \href{https://huggingface.co/meta-llama/Llama-3.1-8B-Instruct}{Link} \\
    & 8B & Tulu3-SFT & \href{https://huggingface.co/allenai/Llama-3.1-Tulu-3-8B-SFT}{Link} \\
    & 8B & Tulu3-DPO & \href{https://huggingface.co/allenai/Llama-3.1-Tulu-3-8B-DPO}{Link} \\
    & 8B & Tulu3-RLVR & \href{https://huggingface.co/allenai/Llama-3.1-Tulu-3-8B}{Link} \\
    \midrule
    Mistral0.3~\citep{jiang2023mistral} & 7B & Instruct & \href{https://huggingface.co/mistralai/Mistral-7B-Instruct-v0.3}{Link} \\
    Falcon3~\citep{almazrouei2023falcon} & 7B & Instruct & \href{https://huggingface.co/tiiuae/Falcon3-7B-Instruct}{Link} \\
    Qwen2.5~\citep{yang2024qwen2_5} & 7B & Instruct & \href{https://huggingface.co/Qwen/Qwen2.5-7B-Instruct}{Link} \\
    Qwen3~\citep{yang2025qwen3} & 32B & Instruct & \href{https://huggingface.co/Qwen/Qwen3-32B}{Link} \\
    \midrule
    \bottomrule
    \end{tabular}
    }
    \caption{The sources of models and tokenizers.}
    \label{tab:app_model_source}
\end{table}

\paragraph{LM Generation.}
By default, we adopt the following key generation settings for LM inference: all the random seeds are 42, the generation temperature is 0 (i.e., deterministic generation without sampling), and the models are loaded in the \texttt{FP16} precision.
For open-weight models containing no more than 8B parameters, the generation is conducted on a single NVIDIA V100 GPU (32GB VRAM).
As for larger models (i.e., Qwen3-32B for distillation fine-tuning), we run each generation session on a single NVIDIA A6000 GPU (50GB VRAM) and apply 4-bit quantization.

\subsection{Fine-tuning Details}
\label{app:finetuning_details}

Each fine-tuning session runs on a single NVIDIA A6000 GPU, with \texttt{unsloth}~\citep{unsloth} employed to accelerate training.
By default, we set the number of epochs to 3, batch size to 8, maximum context length to 4096, model parameter precision to \texttt{BF16}, and all random seeds to 42.
The optimizer is AdamW~\citep{kingma2014adam,loshchilov2017adamw}, with beta1 of 0.9, beta2 of 0.999, epsilon of 1e-8, and maximum gradient norm of 1.
Warmup-Stable-Decay (WSD)~\citep{hu2024minicpm_wsd,wen2025understanding_wsd} is used for learning rate scheduling, where the learning rate linearly increases from 0 to a positive value in the first 5\% training steps, stays the same for 90\% steps, and finally decays to 0 in the remaining 5\% steps with a cosine decaying factor.

In addition, we adopt LoRA~\citep{hu2022lora} fine-tuning and apply grid search on the stable learning rate $\in$ \{2e-5, 5e-5, 2e-5\} and LoRA rank $\in$ \{8, 16, 32\}, with LoRA alpha of $2 \times$ rank and LoRA dropout of 0.

For model selection, we hold out 1\% of the training data as the validation set to evaluate the model checkpoints every 10\% training steps.
During validation, the model uses the same settings as in the LM generation phase, and it is evaluated on 100 randomly sampled instances from the entire validation set.

The final fine-tuned model of a method (such as Raw-FT, DA-FT, or IA-FT) is the one with the best validation score (obtained during fine-tuning) among all the checkpoints across the nine hyperparameter-searching combinations.

\subsection{Prompt Details}
\label{app:prompt_details}

This section presents the baseline methods (Appendix~\ref{app:baseline_methods}), prompt templates for language model evaluation (Appendix~\ref{app:prompt_eval}), prompt templates for model training (Appendix~\ref{app:prompt_train}), and IA prompt variants (Appendix~\ref{app:prompt_var}).

\subsubsection{Baseline Methods}
\label{app:baseline_methods}

As mentioned in \S~\ref{sec:exp_setup_baseline_prompting}, we consider various representative baselines to compare their performance with our IA method.
Here, we present the details of the prompts used in the baseline methods.

\paragraph{DA.}
The basic baseline is DA (Direct Answer), where we do not apply any prompting. In other words, $\mathcal{P}_{\text{DA}}$ is an empty string. Outperforming the DA baseline demonstrates the effectiveness of a prompting method.

\paragraph{CoT.}
CoT (Chain-of-Thought) prompting~\citep{kojima2022cot_think_step_by_step} is a representative reasoning prompting method and widely adopted in model training and inference, which requires LMs to perform step-by-step thinking: $\mathcal{P}_{\text{CoT}}=$ ``\textit{Let's think step by step.}''

\paragraph{PS.}
As a representative planning prompting method, PS (Plan-and-Solve)~\citep{wang2023plan} adopts the following prompting: $\mathcal{P}_{\text{PS}}=$ ``\textit{Let's first understand the problem and devise a plan to solve the problem. Then, let's carry out the plan and solve the problem step by step.}''

\paragraph{AR.}
Utilizing the in-context learning ability~\citep{openai2020gpt3,dong2024icl_survey} of LMs, AR (Analogical Reasoning)~\citep{yasunaga2024analogical} requires the model to recall its memory (i.e., parametric knowledge) before problem-solving: $\mathcal{P}_{\text{AR}}=$ ``\textit{Let's first recall relevant problems as examples. Afterward, proceed to solve the initial problem.}''

\subsubsection{Prompt Templates for Evaluation}
\label{app:prompt_eval}

We use different prompt templates when evaluating language models across various task types.
During evaluation, we disable the system prompt and put the task requirements into the user prompt.
For different prompting methods, we place the prompting message (``\texttt{prompting}'') at the beginning of the assistant prompt to guide the assistant's response (i.e., the model to be evaluated).

\begin{itemize}
\setlength\itemsep{0em}
    \item Figure~\ref{fig:prompt_eval_openqa}: Open-ended question answering (OpenQA) tasks, including TriviaQA and Bamboogle.
    \item Figure~\ref{fig:prompt_eval_mcqa}: Multiple-choice question answering (MCQA) tasks, including MMLU, BBH, and GPQA.
    \item Figure~\ref{fig:prompt_eval_code}: The programming (Coding) task, i.e., MBPP.
\end{itemize}

% \clearpage

% Evaluation prompts

\begin{figure}[t!]
    \centering
    \scalebox{0.98}{
    \begin{tcolorbox}[colback=black!5!white,colframe=black!75!black]
    \normalsize{\textbf{User Prompt}:} \\
    \small{Answer the following question: \\
    \texttt{\{\{question\}\}} } \\
    \\
    \normalsize{\textbf{Assistant Prompt}:} \\
    \small{ \texttt{\{\{prompting\}\}} }
    \end{tcolorbox}
    }
    \caption{Evaluation prompt template for open-ended question answering tasks (without extra context).}
    \label{fig:prompt_eval_openqa}
\end{figure}

\begin{figure}[t!]
    \centering
    \scalebox{0.98}{
    \begin{tcolorbox}[colback=black!5!white,colframe=black!75!black]
    \normalsize{\textbf{User Prompt}:} \\
    \small{Answer the following question by selecting an option: \\
    \texttt{\{\{question\}\}} \\
    \texttt{\{\{options\}\}} } \\
    \\
    \normalsize{\textbf{Assistant Prompt}:} \\
    \small{ \texttt{\{\{prompting\}\}} }
    \end{tcolorbox}
    }
    \caption{Evaluation prompt template for multiple-choice question answering tasks.}
    \label{fig:prompt_eval_mcqa}
\end{figure}

\begin{figure}[t!]
    \centering
    \scalebox{0.98}{
    \begin{tcolorbox}[colback=black!5!white,colframe=black!75!black]
    \normalsize{\textbf{User Prompt}:} \\
    \small{You are a helpful assistant. You are good at Python programming and software development. You can analyze the requirements first and give your final answer at the end. \\
    \\
    Answer Format: Your final answer MUST start with ``Final Answer:'' and the solution MUST be put into a single code block, as in the following example: \\
    \\
    Final Answer: \\
    \texttt{\`{}\`{}\`{}python} \\
    \texttt{<your code here>} \\
    \texttt{\`{}\`{}\`{}} \\
    \\
    Complete the following \texttt{\{\{function\_name\}\}} function. The function header is provided below. Ensure your final answer is grammatically correct, complete, and executable Python code that fulfills the following requirements: \\
    \texttt{\{\{question\}\}} \\
    \\
    \texttt{\{\{function\_header\}\}} } \\
    \\
    \normalsize{\textbf{Assistant Prompt}:} \\
    \small{ \texttt{\{\{prompting\}\}} }
    \end{tcolorbox}
    }
    \caption{Evaluation prompt template for programming tasks.}
    \label{fig:prompt_eval_code}
\end{figure}

\subsubsection{Prompt Templates for Training}
\label{app:prompt_train}

For model fine-tuning, we first ask the labeler model $LM_{\text{label}}$ to analyze the question using the prompt template in Figure~\ref{fig:prompt_train_analysis}.
When the ``\texttt{prompting}'' in Figure~\ref{fig:prompt_train_analysis} is IA prompting (i.e., ``V0'' in Appendix~\ref{app:prompt_var}), the analysis will be used for IA fine-tuning.
Let ``\texttt{analysis}'' be the output of this stage.

Then, we construct the fine-tuning data using the prompt template in Figure~\ref{fig:prompt_train_data}, where ``\texttt{user\_prompt}'' is the same as the user prompts for evaluation (Appendix~\ref{app:prompt_eval}).
For the Raw-FT baseline, both ``\texttt{prompting}'' and ``\texttt{analysis}'' in Figure~\ref{fig:prompt_train_data} are empty.
For the DA-FT baseline, ``\texttt{prompting}'' is empty while ``\texttt{analysis}'' is the problem analysis generated by the Direct Answer method.
For our IA-FT method, ``\texttt{prompting}'' is the IA prompting text and ``\texttt{analysis}'' is the analysis produced by our Intentional Analysis method.

% Training prompts

\begin{figure}[t!]
    \centering
    \scalebox{0.98}{
    \begin{tcolorbox}[colback=black!5!white,colframe=black!75!black]
    \normalsize{\textbf{User Prompt}:} \\
    \small{You will be given a question and the correct answer, and your task is to output the analysis of the question. Your analysis must reasonably lead to the correct answer, but you should not reveal the answer at the beginning. \\
    \\
    Here is the question: \\
    \texttt{\{\{question\}\}} \\
    \\
    We already know the correct answer is \texttt{\{\{answer\}\}} \\
    \\
    Now, you should output the analysis of the question. } \\
    \\
    \normalsize{\textbf{Assistant Prompt}:} \\
    \small{ \texttt{\{\{prompting\}\}} }
    \end{tcolorbox}
    }
    \caption{Training prompt template for question analysis.}
    \label{fig:prompt_train_analysis}
\end{figure}

\begin{figure}[t!]
    \centering
    \scalebox{0.98}{
    \begin{tcolorbox}[colback=black!5!white,colframe=black!75!black]
    \normalsize{\textbf{User Prompt}:} \\
    \small{\texttt{\{\{user\_prompt\}\}} } \\
    \\
    \normalsize{\textbf{Assistant Prompt}:} \\
    \small{ \texttt{\{\{prompting\}\}} \\
    \texttt{\{\{analysis\}\}} \\
    \\ 
    Final Answer: \texttt{\{\{answer\}\}} }
    \end{tcolorbox}
    }
    \caption{Training prompt template for data construction.}
    \label{fig:prompt_train_data}
\end{figure}

% \clearpage

\begin{table*}[t!]
    \centering
    \scalebox{0.75}{
    \begin{tabular}{cl|ccccc|l}
    \toprule
    \midrule
    & \multirow{2}{*}{\textbf{Method}} & General & Knowledge & Compositionality & Reasoning & Coding & \multirow{2}{*}{\textbf{Avg.}} \\
    \cmidrule(lr){3-3} \cmidrule(lr){4-4} \cmidrule(lr){5-5} \cmidrule(lr){6-6} \cmidrule(lr){7-7}
    & & \textbf{MMLU} & \textbf{TriviaQA} & \textbf{Bamboogle} & \textbf{BBH} & \textbf{MBPP} & \\
    \midrule
    \rowcolor{lightgray} & DA (Direct Answer) & 67.71 & 76.93 & 44.80 & 56.02 & 38.72 & 56.84 \\
    \midrule
    \ding{192} & IA Default Prompt (V0) & 68.82 & 79.39 & 47.20 & 60.30 & 51.23 & 61.39 \\
    \ding{193} & IA Prompt Variant (V1) & 68.93 & 79.76 & 48.00 & 62.38 & 47.58 & 61.33 \\
    \ding{194} & IA Prompt Variant (V2) & 69.45 & 79.08 & 47.20 & 61.48 & 49.83 & 61.41 \\
    \ding{195} & IA Prompt Variant (V3) & 69.10 & 79.65 & 48.00 & 61.17 & 48.12 & 61.21 \\
    \ding{196} & IA Prompt Variant (V4) & 69.09 & 79.56 & 45.60 & 60.97 & 48.98 & 60.84 \\
    \ding{197} & IA Prompt Variant (V5) & 69.67 & 79.09 & 45.60 & 60.72 & 47.85 & 60.59 \\
    \midrule
    \multicolumn{2}{c|}{\textit{Avg$_{\text{(Std)}}$ over six IA prompts}} & 69.26$_{(0.4)}$ & 79.42$_{(0.3)}$ & 46.93$_{(1.0)}$ & 61.17$_{(0.7)}$ & 48.93$_{(1.3)}$ & 61.13$_{(0.3)}$ \\
    \midrule
    \bottomrule
    \end{tabular}
    }
    \vspace{-4pt}
    \caption{\textbf{IA Prompt Variants}. The performance (Accuracy\%) of the default IA prompt and its five prompt variants.}
    \label{tab:exp_ia_prompt_var}
    \vspace{-5pt}
\end{table*}

\begin{table*}[ht]
\captionsetup{width=0.48\textwidth}
\centering

\begin{minipage}[t]{0.49\textwidth}
    \centering
    \scalebox{0.6}{
    \begin{tabular}{cc|ccccc|c}
    \toprule
    \midrule
    & \textbf{Temp} & \textbf{MMLU} & \textbf{TriviaQA} & \textbf{Bamboogle} & \textbf{BBH} & \textbf{MBPP} & \textbf{Avg. $\Delta$} \\
    \midrule
    \ding{192} & \textbf{0.25} & \textcolor{teal}{+1.15} & \textcolor{teal}{+2.56} & \textcolor{teal}{+0.80} & \textcolor{teal}{+5.32} & \textcolor{teal}{+8.54} & \textbf{+3.67} \\
    \cmidrule(lr){2-8}
    \ding{193} & \textbf{0.50} & \textcolor{teal}{+1.41} & \textcolor{teal}{+1.73} & \textcolor{teal}{+3.20} & \textcolor{teal}{+4.61} & \textcolor{teal}{+7.84} & \textbf{+3.76} \\
    \cmidrule(lr){2-8}
    \ding{194} & \textbf{0.75} & \textcolor{teal}{+2.02} & \textcolor{teal}{+1.74} & \textcolor{teal}{+13.60} & \textcolor{teal}{+5.21} & \textcolor{teal}{+5.53} & \textbf{+5.62} \\
    \cmidrule(lr){2-8}
    \ding{195} & \textbf{1.00} & \textcolor{teal}{+0.74} & \textcolor{teal}{+0.56} & \textcolor{teal}{+5.60} & \textcolor{teal}{+3.21} & \textcolor{teal}{+4.89} & \textbf{+3.00} \\
    \midrule
    \bottomrule
    \end{tabular}
    }
    \vspace{-4pt}
    \caption{\textbf{Robustness of Intentional Analysis: Generation Temperature}. Regardless of the generation temperature, IA outperforms DA consistently. Here, we fix the random seeds to 42 and Top-p sampling to 0.9.}
    \label{tab:exp_gen_temp}
\end{minipage}
\hfill
\begin{minipage}[t]{0.49\textwidth}
    \centering
    \scalebox{0.6}{
    \begin{tabular}{cc|ccccc|c}
    \toprule
    \midrule
    & \textbf{Seeds} & \textbf{MMLU} & \textbf{TriviaQA} & \textbf{Bamboogle} & \textbf{BBH} & \textbf{MBPP} & \textbf{Avg. $\Delta$} \\
    \midrule
    \ding{192} & \textbf{7} & \textcolor{teal}{+1.88} & \textcolor{teal}{+1.78} & \textcolor{teal}{+5.60} & \textcolor{teal}{+3.85} & \textcolor{teal}{+3.97} & \textbf{+3.42} \\
    \cmidrule(lr){2-8}
    \ding{193} & \textbf{42} & \textcolor{teal}{+1.58} & \textcolor{teal}{+0.89} & \textcolor{teal}{+4.00} & \textcolor{teal}{+5.12} & \textcolor{teal}{+5.87} & \textbf{+3.49} \\
    \cmidrule(lr){2-8}
    \ding{194} & \textbf{365} & \textcolor{teal}{+2.24} & \textcolor{teal}{+2.03} & \textcolor{teal}{+4.80} & \textcolor{teal}{+5.48} & \textcolor{teal}{+8.37} & \textbf{+4.58} \\
    \cmidrule(lr){2-8}
    \ding{195} & \textbf{1024} & \textcolor{teal}{+1.74} & \textcolor{teal}{+2.23} & \textcolor{teal}{+5.60} & \textcolor{teal}{+4.87} & \textcolor{teal}{+4.95} & \textbf{+3.88} \\
    \midrule
    \bottomrule
    \end{tabular}
    }
    \vspace{-4pt}
    \caption{\textbf{Robustness of Intentional Analysis: Random Seeds}. Irrespective of the random seeds, IA brings consistent improvements. Here, we fix the generation temperature to 0.7 and Top-p sampling to 0.9.}
    \label{tab:exp_random_seed}
\end{minipage}
\vspace{-5pt}
\end{table*}

\begin{table*}[t!]
    \centering
    \scalebox{0.75}{
    \begin{tabular}{cl|ccccc|l}
    \toprule
    \midrule
    & \multirow{2}{*}{\textbf{Method}} & General & Knowledge & Compositionality & Reasoning & Coding & \multirow{2}{*}{\textbf{Avg.}} \\
    \cmidrule(lr){3-3} \cmidrule(lr){4-4} \cmidrule(lr){5-5} \cmidrule(lr){6-6} \cmidrule(lr){7-7}
    & & \textbf{MMLU} & \textbf{TriviaQA} & \textbf{Bamboogle} & \textbf{BBH} & \textbf{MBPP} & \\
    \midrule
    \rowcolor{lightgray}\ding{192} & DA (Direct Answer) & 53.64 & 74.58 & \underline{33.00} & \underline{37.47} & 35.33 & 46.80 \\
    \rowcolor{ia_yellow}\ding{194} & \textbf{IA} (Intentional Analysis) & \underline{55.61} & \textbf{75.40} & \textbf{35.20} & \textbf{41.63} & \textbf{42.57} & \textbf{50.08} \\
    \ding{193} & MI (Mistral-Interact) & 55.42 & 69.32 & 26.40 & 27.98 & 35.78 & 42.98 \\
    \ding{195} & MI+IA & \textbf{55.94} & \underline{75.02} & 30.40 & 36.16 & \underline{38.67} & \underline{47.24} \\
    \midrule
    \bottomrule
    \end{tabular}
    }
    \vspace{-4pt}
    \caption{\textbf{Additional Comparisons}. The performance (Accuracy\%) of our IA method, the DA baseline, the MI model, and the combined method MI+IA on representative benchmarks requiring diverse capabilities.}
    \label{tab:exp_ia_vs_mi}
    \vspace{-5pt}
\end{table*}

\section{Further Analysis}
\label{app:further_analysis}

\subsection{IA Prompt Variants}
\label{app:prompt_var}

As mentioned in \S~\ref{sec:results_prompting_effectiveness}, to demonstrate that IA works effectively irrespective of specific prompt design, we conduct experiments on different IA prompt variants.
The default IA prompt (``V0'') is paraphrased into five different versions as follows:
\begin{itemize}
\setlength\itemsep{0em}
    \item V0: ``\textit{Let's analyze the intent of the question and then answer.}''
    \item V1: ``\textit{Let's identify the question's intent and then work through a logical solution.}''
    \item V2: ``\textit{Let's examine the intention of the question and then provide a well-reasoned answer.}''
    \item V3: ``\textit{Let's interpret the intention behind the question and then proceed to solve it.}''
    \item V4: ``\textit{Let's understand the intention of the question and then reason through the answer.}''
    \item V5: ``\textit{Let's clarify the question's intent and then build the answer using logical reasoning.}''
\end{itemize}

Table~\ref{tab:exp_ia_prompt_var} presents the performance of the DA baseline and our IA method with different IA prompts, using the LLaMA3-8B-Instruct model.
The consistency of the variants' superior performance verify that IA works effectively regardless of its specific prompt design, so long as the model faithfully performs intentional analysis as instructed.

\subsection{Robustness of Intentional Analysis}
\label{app:results_prompting_robustness}

To explore the robustness of our IA method under varied LM generation settings, we investigate the performance perturbations when tuning the key hyperparameters controlling generation randomness: the temperature and random seeds.
Table~\ref{tab:exp_gen_temp} and Table~\ref{tab:exp_random_seed} report the performance gains (IA over DA) on diverse benchmarks using the LLaMA3-8B-Instruct model.
In Table~\ref{tab:exp_gen_temp}, we try different temperatures when fixing the random seeds to 42;
In Table~\ref{tab:exp_random_seed}, we try different random seeds when fixing the temperature to 0.7, i.e., not deterministic generation.
In both tables, the Top-p sampling is set to 0.9.
The results of both tables show that IA brings consistent gains under varying generation configurations, verifying the robustness of our IA method.

\subsection{Additional Comparisons}
\label{app:exp_ia_vs_mi}

Here, we compare our IA method with \textbf{MI} (Mistral-Interact)~\citep{qian2024in3_mi}, a fine-tuned Mistral-7B~\citep{jiang2023mistral} model on the Intention-in-Interaction (IN3) benchmark, which inspects users' implicit intentions through explicit queries.

Table~\ref{tab:exp_ia_vs_mi} shows the performance of our IA method, the DA baseline, the MI model, and the combined method MI+IA (i.e., applying IA to the MI model), where DA and IA are using the Mistral0.3-7B-Instruct model.
We observe that IA outperforms MI on all five benchmarks, possibly because IA is a general-purpose problem-solving method empowered by intentional analysis, while the training data of the MI model primarily focuses on user-agent conversations rather than problem-solving or reasoning.

In addition, the fact that MI+IA is consistently better than MI further solidifies the effectiveness of IA for general problem-solving using various language models (i.e., either original Mistral model or fine-tuned Mistral-Interact model).

\begin{table}[t!]
    \centering
    \scalebox{0.75}{
    \begin{tabular}{ccccccc}
    \toprule
    \midrule
    \multirow{2}{*}{\textbf{Dataset}} & \multicolumn{3}{c}{\textbf{\# Input Tokens}} & \multicolumn{3}{c}{\textbf{\# Output Tokens}} \\
    \cmidrule(lr){2-4} \cmidrule(lr){5-7}
    & DA & IA & $\Delta$ & DA & IA & $\Delta$ \\
    \midrule
    MMLU & 147 & 159 & +12 & 168 & 201 & +20\% \\ % +33
    TriviaQA & 59 & 71 & +12 & 58 & 71 & +22\% \\ % +13
    Bamboogle & 54 & 66 & +12 & 121 & 81 & -33\% \\ % -40
    BBH & 161 & 173 & +12 & 128 & 152 & +19\% \\ % +24
    MBPP & 183 & 195 & +12 & 210 & 428 & +104\% \\ % +218
    \midrule
    \bottomrule
    \end{tabular}
    }
    \vspace{-4pt}
    \caption{\textbf{Efficiency Study.} The number of input and output tokens of the DA baseline and our IA method.}
    \label{tab:stat_efficiency}
    \vspace{-12pt}
\end{table}

\subsection{Efficiency Study}
\label{app:efficiency_study}

To study the efficiency cost of requiring LMs to conduct intentional analysis, we present the number of input and output tokens of the DA baseline and our IA method in Table~\ref{tab:stat_efficiency}.
IA adds $12$ extra input tokens due to the IA prompting (i.e., ``V0'' in Appendix~\ref{app:prompt_var}) for the additional triggering instruction, and the number of extra output tokens is normally about $20\%$. The IA solutions for the Bamboogle task are more concise, and those for the MBPP coding task are more lengthy.
Hence, this cost should be traded off against the substantially improved performance of the proposed IA method.

\begin{figure*}[t!]
\captionsetup{width=0.98\textwidth}
\centering
\begin{minipage}[t]{0.64\textwidth}
    \centering
    % \vspace{-5pt}
    \includegraphics[width=0.98\textwidth]{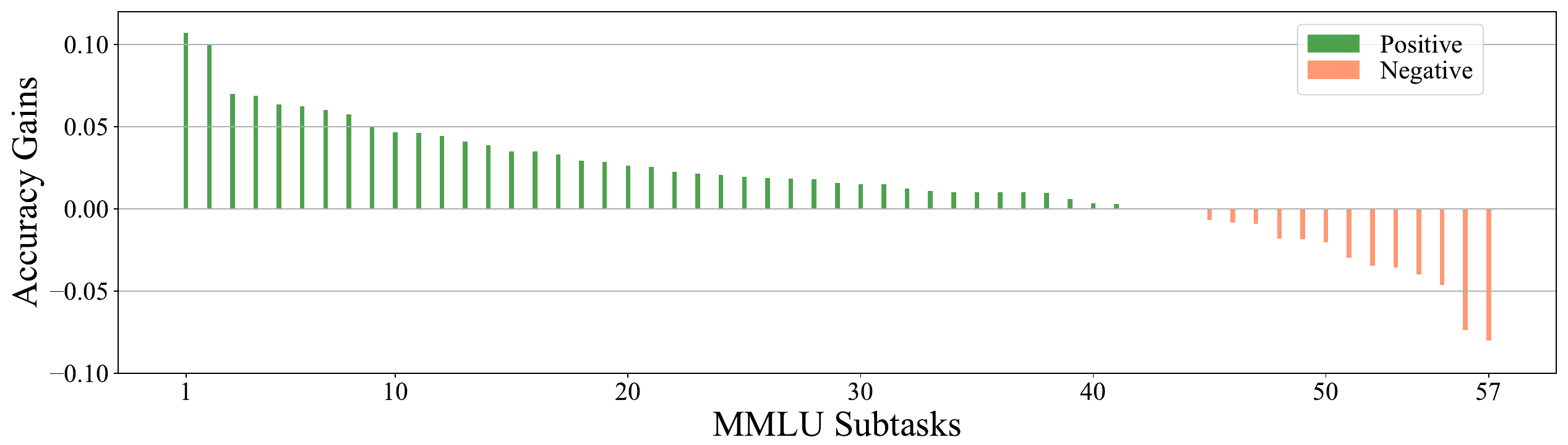}
    % \vspace{-3pt}
    % \caption{IA performance on different MMLU subtasks.}
    % \label{fig:ia_subtask_mmlu}
    % \vspace{-5pt}
\end{minipage}
\hfill
\begin{minipage}[t]{0.34\textwidth}
    \centering
    % \vspace{-5pt}
    \includegraphics[width=0.98\linewidth]{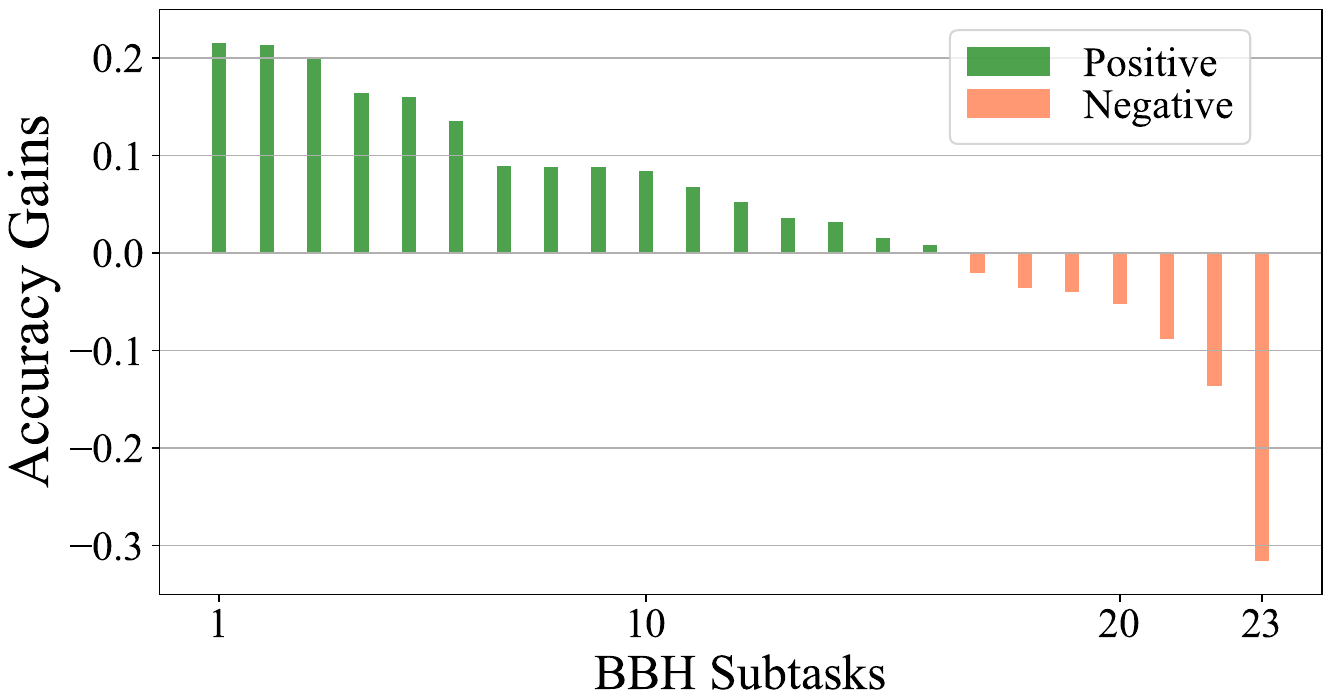}
    % \vspace{-3pt}
    % \caption{IA performance on different BBH subtasks.}
    % \label{fig:ia_subtask_bbh}
    % \vspace{-5pt}
\end{minipage}
\vspace{-3pt}
\caption{Performance gains by IA over DA on different subtasks from MMLU and BBH benchmarks.}
\label{fig:ia_subtask}
\vspace{-5pt}
\end{figure*}

\subsection{Scenarios in which IA is more beneficial}
\label{app:ia_subtask}

We observe a consistent improvement from our IA method on five distinct task categories in Table~\ref{tab:exp_ia_effectiveness}.
Among the five benchmarks, MMLU (General type) and BBH (Reasoning type) have fine-grained subdivisions, covering various kinds of subtasks that require different skill sets.
Hence, we are interested in which tasks IA would be more beneficial for.
Figure~\ref{fig:ia_subtask} presents the performance gains by our IA method over the DA baseline across all the MMLU subtasks and BBH subtasks.

% 57 = 41 + 3 + 13
On the MMLU benchmark, IA outperforms DA on 41 subtasks (72\%). 
Top 8 subtasks for which IA achieves a performance gain of more than 5 points include ``\texttt{machine learning}'', ``\texttt{global facts}'', ``\texttt{abstract algebra}'', ``\texttt{human sexuality}'', ``\texttt{elementary mathematics}'', ``\texttt{college biology}'', ``\texttt{business ethics}'', and ``\texttt{miscellaneous}''.
For the reasoning benchmark BBH, IA surpasses DA on 16 subtasks (70\%), where IA brings an accuracy gain of above 10 points on six tasks, including ``\texttt{logical deduction three objects}'', ``\texttt{snarks}'', ``\texttt{date understanding}'', ``\texttt{reasoning about colored objects}'', ``\texttt{temporal sequences}'', and ``\texttt{disambiguation qa}''.
% 23 = 16 + 7
On the other hand, IA underperforms DA on 13 subtasks of MMLU, with a performance degradation of over 5 points on ``\texttt{moral scenarios}'' and ``\texttt{us foreign policy}''.
For BBH, IA performs unsatisfactorily on ``\texttt{web of lies}'' and ``\texttt{causal judgment}'', where we observe a performance drop of above 10 points.

Regarding specific scenarios where IA performs unsatisfactorily, we look into the failure cases and find a common property of the tasks that IA fails more on:
The ``Moral Scenarios'' subtask of MMLU gives two scenarios (short stories), and the model is asked to judge if each scenario is morally problematic or not. Here, the question is ``\textit{For which of these two scenarios does the main character do something clearly morally wrong?}''. Similarly, the ``Causal Judgment'' subtask of BBH gives a short story of a person and asks ``\textit{How would a typical person answer each of the following questions about causation?}''.
Therefore, we hypothesize that, in such tasks, analyzing the character's intent in the story should be more crucial than analyzing the question itself.
To verify this hypothesis, we conducted a proof-of-concept experiment, where we prompt the model to \textit{analyze the intent of the character in the given story and then answer}, and this change leads to a performance gain of +3 points on the aforementioned task. This suggests an interesting direction of future work on hierarchical intent analysis.

Overall, these findings provide useful clues regarding the suitable tasks and scenarios for applying IA and suggest research directions for future work on improving LMs with intentional analysis.

\subsection{Examples in Case Studies}
\label{app:case_study}

To provide further insights into the effect of the proposed IA method, we conduct case studies on random samples from the evaluation results in Table~\ref{tab:exp_ia_effectiveness}.
Here, Table~\ref{tab:case_study_mmlu}, Table~\ref{tab:case_study_trivia_qa}, Table~\ref{tab:case_study_bamboogle}, Table~\ref{tab:case_study_bbh}, and Table~\ref{tab:case_study_mbpp} present concrete cases from MMLU, TriviaQA, Bamboogle, BBH, and MBPP benchmarks, respectively, where our IA method correctly answers the question but the DA baseline fails.

As analyzed in \S~\ref{sec:how_ia_vs_da}, these case studies reveal problems in the baseline method, such as \textit{intent misunderstanding}, \textit{hasty generalization}, \textit{mental laziness}, and \textit{faulty reasoning}.
Instead, our IA method appears to mitigate the issues with its intentional solution.

\clearpage
\begin{table*}[t!]
    \centering
    \scalebox{0.6}{
    \begin{tabular}{ccl}
    \toprule
    \midrule
    \ding{192} & \textbf{Question} & \makecell[l]{On Andros Islands in the Bahamas, populations of mosquito fish, Gambusia hubbsi, colonized a series of ponds. These ponds are \\ no longer connected. However, the environments are very similar except that some ponds contain predatory fish, while others do not. \\ In high predation ponds, selection has favored the evolution of a body shape that enables rapid bursts of speed. \\ In low predation ponds, another body type is favored--one that is well-suited to swim for long periods of time. \\ (A) Reproductive isolation caused geographic isolation. \\ (B) Reproductive isolation was not complete. \\ \textcolor{teal}{(C) Allopatric isolation brought about reproductive isolation.} \\ (D) Sympatric isolation brought about reproductive isolation.} \\
    \cmidrule(lr){2-3}
    & \textbf{Answer} & (C) \\
    \cmidrule(lr){2-3}
    & \textbf{DA} output: & \makecell[l]{\textcolor{red}{The correct answer is (D)} Sympatric isolation brought about reproductive isolation. \\ \textcolor{red}{Explanation:} Sympatric speciation occurs when a new species emerges from a single ancestral species in the same geographic area, \\ often due to ecological differences. In this case, the mosquito fish populations in the ponds with and without predators \\ have adapted to their respective environments, leading to the emergence of two distinct body types. This is an example of \\ sympatric speciation, where the reproductive isolation is not due to geographic barriers but rather ecological differences.} \\
    \cmidrule(lr){2-3}
    & \textbf{IA} output: & \makecell[l]{\textcolor{blue}{The question is asking about} the mechanism of reproductive isolation in the context of the evolution of the mosquito fish populations \\ on the Andros Islands. \textcolor{blue}{Reproductive isolation is} the process by which a new species emerges due to the inability of individuals from \\ different populations to interbreed. \textcolor{blue}{The key information in the question is} that the ponds are ``no longer connected.'' \\ \textcolor{blue}{This implies that} the mosquito fish populations in the different ponds are geographically isolated from one another. \\ \textcolor{blue}{Now, let's look at the options}: \\ (A) Reproductive isolation caused geographic isolation: This option is incorrect because geographic isolation is the result of \\ \quad \quad reproductive isolation, not the cause. \\ (B) Reproductive isolation was not complete: \textcolor{blue}{This option is not relevant to the question, which is asking about} \\ \quad \quad the mechanism of reproductive isolation. \\ (C) Allopatric isolation brought about reproductive isolation: Allopatric isolation is a type of geographic isolation \\ \quad \quad where a new species emerges due to the physical separation of a population from the rest of the species. \\ \quad \quad This option is correct because the mosquito fish populations in the different ponds are geographically isolated from one another. \\ (D) Sympatric isolation brought about reproductive isolation: Sympatric isolation is a type of reproductive isolation \\ \quad \quad where a new species emerges without geographic separation. This option is incorrect because \\ \quad \quad the mosquito fish populations in the different ponds are geographically isolated from one another. \\ \textcolor{teal}{The correct answer is (C)} Allopatric isolation brought about reproductive isolation.} \\
    \midrule
    \ding{193} & \textbf{Question} & \makecell[l]{Which of the following poses the greatest safety risk while being heated in a school laboratory? \\ (A) A mixture of iron and sulfur \quad \textcolor{teal}{(B) Mercury(II) oxide} \quad (C) Sodium chloride \quad (D) Copper(II) sulfate hydrate} \\
    \cmidrule(lr){2-3}
    & \textbf{Answer} & (B) \\
    \cmidrule(lr){2-3}
    & \textbf{DA} output: & \makecell[l]{\textcolor{red}{The correct answer is (A)} A mixture of iron and sulfur. When iron and sulfur are heated together, \\ they can undergo a highly exothermic reaction, producing iron(II) sulfide and releasing a significant amount of heat. \\ This reaction can lead to a rapid increase in temperature, potentially causing a fire or explosion. \\ This reaction is known as a thermite reaction, and it is often used in pyrotechnics and welding. \\ The other options do not pose the same level of safety risk when heated in a school laboratory: \\ \quad - Mercury(II) oxide (B) is a relatively stable compound and does not pose a significant safety risk when heated. \\ \quad - Sodium chloride (C) is a stable compound and does not pose a significant safety risk when heated. \\ \quad - Copper(II) sulfate hydrate (D) can decompose when heated, producing a gas, but it is not typically considered \\ \quad \quad \quad a high-risk substance in a laboratory setting.} \\
    \cmidrule(lr){2-3}
    & \textbf{IA} output: & \makecell[l]{\textcolor{blue}{The question is asking about} the substance that poses the greatest safety risk when heated in a school laboratory. \\ \textcolor{blue}{This means we are looking for} a substance that is highly reactive, potentially explosive, or releases toxic gases when heated. \\ (A) A mixture of iron and sulfur: When heated, iron and sulfur can form iron(II) sulfide, which is not particularly hazardous. \\ (B) Mercury(II) oxide: When heated, mercury(II) oxide decomposes to release mercury vapor, which is toxic and \\ \quad \quad can be hazardous if inhaled. \\ (C) Sodium chloride: When heated, sodium chloride (common table salt) does not pose a significant safety risk. \\ (D) Copper(II) sulfate hydrate: When heated, copper(II) sulfate hydrate decomposes to release sulfur trioxide gas, which is toxic, \\ \quad \quad but the risk is not as high as that of mercury vapor. \\ \textcolor{blue}{Considering the options}, mercury(II) oxide poses the greatest safety risk when heated in a school laboratory \\ \quad \quad due to the release of toxic mercury vapor.} \\
    \midrule
    \ding{194} & \textbf{Question} & \makecell[l]{Which planet(s) in the solar system has/have active plate tectonics? \\ (A) Mars \quad (B) Venus \quad \textcolor{teal}{(C) Earth} \quad (D) Mars and Earth} \\
    \cmidrule(lr){2-3}
    & \textbf{Answer} & (C) \\
    \cmidrule(lr){2-3}
    & \textbf{DA} output: & \makecell[l]{\textcolor{red}{The correct answer is (D)} Mars and Earth. \\ Mars has evidence of past tectonic activity, with features such as volcanoes, canyons, and evidence of water flow. \\ While it is not as active as Earth, Mars is believed to have experienced tectonic activity in the past. \\ Earth is the only known planet in the solar system with active plate tectonics, where the lithosphere is broken into several \\ large plates that move relative to each other, resulting in earthquakes, volcanic activity, and the creation of mountain ranges. \\ Venus, on the other hand, is thought to have experienced tectonic activity in the past, but it is now considered a \\ geologically inactive planet, with a surface that is relatively young and smooth, indicating a lack of recent tectonic activity.} \\
    \cmidrule(lr){2-3}
    & \textbf{IA} output: & \makecell[l]{\textcolor{blue}{The question is asking about} active plate tectonics, \textcolor{blue}{which refers to} the movement of the Earth's lithosphere \\ (the outermost solid layer of the planet) and the processes that shape the planet's surface, \\ such as earthquakes, volcanic eruptions, and the creation of mountain ranges. \\ Option (A) Mars: Mars has evidence of past tectonic activity, but it is not considered to have active plate tectonics. \\ \quad \quad Mars' surface is relatively quiet, with few signs of recent tectonic activity. \\ Option (B) Venus: Venus has a very slow rate of tectonic activity, but it is not considered to have active plate tectonics. \\ \quad \quad Venus' surface is characterized by numerous volcanoes and lava flows, \\ \quad \quad but the rate of volcanic activity is much lower than on Earth. \\ Option (C) Earth: Earth is the only planet in the solar system with active plate tectonics. The movement of the Earth's \\ \quad \quad lithosphere is responsible for the creation of mountain ranges, earthquakes, and volcanic eruptions. \\ Option (D) Mars and Earth: While both planets have evidence of past tectonic activity, neither is considered to have \\ \quad \quad active plate tectonics. \\ \textcolor{teal}{The correct answer is (C)} Earth.} \\
    \midrule
    \bottomrule
    \end{tabular}
    }
    \caption{\textbf{Cases from MMLU results}, where IA correctly answers the question while the DA baseline fails. The DA baseline often rushes to conclusions and then tries to present a post-hoc explanation, while our IA method begins with restating the question and detailing the key points, followed by point-by-point analysis.}
    \label{tab:case_study_mmlu}
\end{table*}

\clearpage
\begin{table*}[t!]
    \centering
    \scalebox{0.6}{
    \begin{tabular}{ccl}
    \toprule
    \midrule
    \ding{192} & \textbf{Question} & \makecell[l]{Which line on the London Underground rail system has the most stations?} \\
    \cmidrule(lr){2-3}
    & \textbf{Answer} & District Line \\
    \cmidrule(lr){2-3}
    & \textbf{DA} output: & \makecell[l]{\textcolor{red}{The Central line} on the London Underground rail system has the most stations, with a total of 49 stations.} \\
    \cmidrule(lr){2-3}
    & \textbf{IA} output: & \makecell[l]{\textcolor{blue}{The question is asking for} the line with the most stations on the London Underground rail system. \\ The answer is the \textcolor{teal}{District line}, which has 60 stations.} \\
    \midrule
    \ding{193} & \textbf{Question} & \makecell[l]{British band Lindisfarne had a 1971 hit album entitled ``The Fog on the where''?} \\
    \cmidrule(lr){2-3}
    & \textbf{Answer} & Tyne \\
    \cmidrule(lr){2-3}
    & \textbf{DA} output: & \makecell[l]{\textcolor{red}{The answer is ``the Mountain''.} The album is ``Fog on the Mountain''.} \\
    \cmidrule(lr){2-3}
    & \textbf{IA} output: & \makecell[l]{\textcolor{blue}{The question is asking for} the title of a 1971 album by the British band Lindisfarne, \\ which includes the phrase ``The Fog on the where''. \textcolor{teal}{The correct answer is ``Fog on the Tyne''.}} \\
    \midrule
    \ding{194} & \textbf{Question} & \makecell[l]{``The Battle of the Ages'' took place in April 1991 between George Foreman and which other boxer?} \\
    \cmidrule(lr){2-3}
    & \textbf{Answer} & Evander Holyfield \\
    \cmidrule(lr){2-3}
    & \textbf{DA} output: & \makecell[l]{\textcolor{red}{I'm not aware of any information about} a boxing match called ``The Battle of the Ages'' between George Foreman \\ and another boxer in April 1991.} \\
    \cmidrule(lr){2-3}
    & \textbf{IA} output: & \makecell[l]{\textcolor{blue}{The question is asking for} a boxer who fought George Foreman in a notable fight, and the fight is referred to as \\ ``The Battle of the Ages''. The most notable fight between George Foreman and another boxer that fits this description \\ is the ``Battle of the Ages'' fight between George Foreman and \textcolor{teal}{Evander Holyfield}.} \\
    \midrule
    \ding{195} & \textbf{Question} & \makecell[l]{Where has Baroness Royall of Blaisdon led the opposition since May 2010?} \\
    \cmidrule(lr){2-3}
    & \textbf{Answer} & House of Lords \\
    \cmidrule(lr){2-3}
    & \textbf{DA} output: & \makecell[l]{\textcolor{red}{I do not have information on} where Baroness Royall of Blaisdon has led the opposition since May 2010.} \\
    \cmidrule(lr){2-3}
    & \textbf{IA} output: & \makecell[l]{\textcolor{blue}{The question is asking about} a specific person, Baroness Royall of Blaisdon, and her role in opposition since May 2010. \\ Baroness Royall of Blaisdon is a British Labour Party politician. She was the Leader of the Opposition in the \textcolor{teal}{House of Lords} \\ from 2010 to 2016.} \\
    \midrule
    \bottomrule
    \end{tabular}
    }
    \caption{\textbf{Cases from TriviaQA results}, where IA correctly answers the question while the DA baseline fails. The intentional analysis stage of IA helps the model better retrieve its own parametric knowledge, instead of recalling the incorrect memory or giving up trying. Thus, the model will not simply say ``I don't know'' even though it actually knows the correct answer.}
    \label{tab:case_study_trivia_qa}
\end{table*}

% \clearpage
\begin{table*}[t!]
    \centering
    \scalebox{0.6}{
    \begin{tabular}{ccl}
    \toprule
    \midrule
    \ding{192} & \textbf{Question} & \makecell[l]{Which Theranos whistleblower is not related to a senior American government official?} \\
    \cmidrule(lr){2-3}
    & \textbf{Answer} & Erika Cheung \\
    \cmidrule(lr){2-3}
    & \textbf{DA} output: & \makecell[l]{\textcolor{red}{I couldn't find any information} that suggests a Theranos whistleblower is related to a senior American government official.} \\
    \cmidrule(lr){2-3}
    & \textbf{IA} output: & \makecell[l]{\textcolor{blue}{The question is asking for} a Theranos whistleblower who is not related to a senior American government official. \\ Erika Cheung and Tyler Shultz were two whistleblowers at Theranos. \\ \textcolor{teal}{Erika Cheung} is not related to a senior American government official.} \\
    \midrule
    \ding{193} & \textbf{Question} & \makecell[l]{In what year was the company that was founded as Sound of Music added to the S\&P 500?} \\
    \cmidrule(lr){2-3}
    & \textbf{Answer} & 1999 \\
    \cmidrule(lr){2-3}
    & \textbf{DA} output: & \makecell[l]{\textcolor{red}{I am unable to verify} when the company that was founded as Sound of Music was added to the S\&P 500.} \\
    \cmidrule(lr){2-3}
    & \textbf{IA} output: & \makecell[l]{The company that was founded as Sound of Music \textcolor{blue}{is likely referring to} the company that was originally called Sound of Music, \\ but later changed its name to Panera Bread. Therefore, \textcolor{teal}{the answer is 1999}.} \\
    \midrule
    \bottomrule
    \end{tabular}
    }
    \caption{\textbf{Cases from Bamboogle results}, where IA correctly answers the question while the DA baseline fails.}  % \textbf{Case study.}
    \label{tab:case_study_bamboogle}
\end{table*}

\clearpage
\begin{table*}[t!]
    \centering
    \scalebox{0.6}{
    \begin{tabular}{ccl}
    \toprule
    \midrule
    \ding{192} & \textbf{Problem} & \makecell[l]{On the desk, there is an orange scrunchiephone charger, a turquoise bracelet, a gold sheet of paper, and a red envelope. \\ Is the bracelet turquoise? \\ Options: \textcolor{teal}{(A) yes} \quad (B) no} \\
    \cmidrule(lr){2-3}
    & \textbf{Answer} & (A) \\
    \cmidrule(lr){2-3}
    & \textbf{DA} output: & \makecell[l]{\textcolor{red}{The correct answer is (B) no.}} \\
    \cmidrule(lr){2-3}
    & \textbf{IA} output: & \makecell[l]{\textcolor{blue}{The question is asking} if the bracelet is turquoise. \textcolor{blue}{The information given states that} there is a turquoise bracelet on the desk. \\ \textcolor{teal}{The correct answer is (A) yes.}} \\
    \midrule
    \ding{193} & \textbf{Problem} & \makecell[l]{Find a movie similar to Batman, The Mask, The Lion King, Pretty Woman: \\ Options: \\ (A) Los Angeles Plays Itself \quad (B) Gintama The Final Chapter - Be Forever Yorozuya \\ (C) Doctor Who Last Christmas \quad \textcolor{teal}{(D) The Fugitive}} \\
    \cmidrule(lr){2-3}
    & \textbf{Answer} & (D) \\
    \cmidrule(lr){2-3}
    & \textbf{DA} output: & \makecell[l]{\textcolor{red}{I cannot verify} which movie is similar to Batman, The Mask, The Lion King, and Pretty Woman.} \\
    \cmidrule(lr){2-3}
    & \textbf{IA} output: & \makecell[l]{\textcolor{blue}{The question is asking for} a movie similar to Batman, The Mask, The Lion King, and Pretty Woman. These movies are \\ all action-packed, have a mix of drama and comedy, and often feature a transformation or a change in the main character. \\ \textcolor{blue}{Considering the options provided}: \\ (A) Los Angeles Plays Itself - This is a documentary film that explores the city of Los Angeles, \\ \quad \quad which doesn't fit the action-packed and transformation theme of the given movies. \\ (B) Gintama The Final Chapter - Be Forever Yorozuya - This is an anime film that is action-packed and has a mix of \\ \quad \quad drama and comedy, but it's not similar to the other movies mentioned in terms of its genre or style. \\ (C) Doctor Who Last Christmas - This is a science fiction episode of the TV series Doctor Who, \\ \quad \quad which doesn't fit the theme of the given movies. \\ (D) The Fugitive - This is an action-thriller film that features a transformation of the main character, \\ \quad \quad who is on the run and must clear his name. This fits the theme of the given movies. \\ Therefore, \textcolor{teal}{the best answer is (D)} The Fugitive.} \\
    \midrule
    \ding{194} & \textbf{Problem} & \makecell[l]{The following paragraphs each describe a set of three objects arranged in a fixed order. The statements are logically \\ consistent within each paragraph. \\ In a golf tournament, there were three golfers: Amy, Ada, and Joe. Joe finished below Ada. Amy finished above Ada. \\ Options: (A) Amy finished second \quad \textcolor{teal}{(B) Ada finished second} \quad (C) Joe finished second} \\
    \cmidrule(lr){2-3}
    & \textbf{Answer} & (B) \\
    \cmidrule(lr){2-3}
    & \textbf{DA} output: & \makecell[l]{To solve this, let's analyze the information given: \\ \quad 1. Joe finished below Ada: This means Joe finished either second or third, and Ada finished above him (either first or second). \\ \quad 2. Amy finished above Ada: This means Amy finished either first or second, and Ada finished below her (either second or third). \\ Combining the two pieces of information, we can conclude that \textcolor{teal}{Amy finished first}, Ada finished second, and Joe finished third. \\ So, \textcolor{red}{the correct answer is: (A) Amy finished second}} \\
    \cmidrule(lr){2-3}
    & \textbf{IA} output: & \makecell[l]{\textcolor{blue}{The question is asking for} the position of Ada in the tournament. \\ Since Amy finished above Ada, and Joe finished below Ada, we can conclude that \textcolor{teal}{Ada must be in the middle}. \\ So, \textcolor{teal}{the correct answer is (B) Ada finished second.}} \\
    \midrule
    \bottomrule
    \end{tabular}
    }
    \caption{\textbf{Cases from BBH results}, where IA correctly answers the question while the DA baseline fails. The intentional analysis at the beginning helps with the subsequent reasoning by specifying the correct direction.}
    \label{tab:case_study_bbh}
\end{table*}

\clearpage
\begin{table*}[t!]
    \centering
    \scalebox{0.6}{
    \begin{tabular}{ccl}
    \toprule
    \midrule
    \ding{192} & \textbf{Problem} & \makecell[l]{Write a function to perform the exponentiation of the given two tuples. \\ \texttt{def find\_exponentio(test\_tup1, test\_tup2):}} \\
    \cmidrule(lr){2-3}
    & \textbf{Answer} & \makecell[l]{\texttt{def find\_exponentio(test\_tup1, test\_tup2):} \\ \quad \quad \texttt{res = tuple(ele1 ** ele2 for ele1, ele2 in zip(test\_tup1, test\_tup2))} \\ \quad \quad \texttt{return (res)}} \\
    \cmidrule(lr){2-3}
    & \textbf{DA} output: & \makecell[l]{Final Answer: \\ \textcolor{red}{$\dots$ (buggy code)}} \\
    \cmidrule(lr){2-3}
    & \textbf{IA} output: & \makecell[l]{The function \texttt{\`{}find\_exponentio\`{}} is supposed to perform exponentiation of two tuples. \\ However, exponentiation is typically a mathematical operation between two numbers, not tuples. \\ \textcolor{blue}{Assuming the intention is to} perform element-wise exponentiation of the corresponding elements in the two tuples, \\ we can proceed with the solution. Here's the solution: \\ \texttt{\`{}\`{}\`{}python} \\ \texttt{def find\_exponentio(test\_tup1, test\_tup2):} \\ \quad \quad \texttt{"""} \\ \quad \quad \texttt{This function performs element-wise exponentiation of two tuples.} \\ \quad \quad \texttt{Args:} \\ \quad \quad \quad \quad \texttt{test\_tup1 (tuple): The base tuple.} \\ \quad \quad \quad \quad \texttt{test\_tup2 (tuple): The exponent tuple.} \\ \quad \quad \texttt{Returns:} \\ \quad \quad \quad \quad \texttt{tuple: A new tuple with the result of element-wise exponentiation.} \\ \quad \quad \texttt{"""} \\ \quad \quad \texttt{\# Check if both inputs are tuples} \\ \quad \quad \texttt{if not isinstance(test\_tup1, tuple) or not isinstance(test\_tup2, tuple):} \\ \quad \quad \quad \quad \texttt{raise TypeError("Both inputs must be tuples.")} \\ \\ \quad \quad \texttt{\# Check if both tuples have the same length} \\ \quad \quad \texttt{if len(test\_tup1)!= len(test\_tup2):} \\ \quad \quad \quad \quad \texttt{raise ValueError("Both tuples must have the same length.")} \\ \\ \quad \quad \texttt{\# Perform element-wise exponentiation} \\ \quad \quad \textcolor{teal}{\texttt{result = tuple(base ** exponent for base, exponent in zip(test\_tup1, test\_tup2))}} \\ \\ \quad \quad \texttt{return result} \\ \texttt{\`{}\`{}\`{}} } \\
    \midrule
    \ding{193} & \textbf{Problem} & \makecell[l]{Write a python function to find the hamming distance between given two integers. \\ \texttt{def hamming\_distance(n1, n2):}} \\
    \cmidrule(lr){2-3}
    & \textbf{DA} output: & \makecell[l]{Final Answer: \\ \textcolor{red}{$\dots$ (buggy code)}} \\
    \cmidrule(lr){2-3}
    & \textbf{IA} output: & \makecell[l]{The \textcolor{blue}{Hamming distance between two integers is} the number of positions at which the corresponding bits are different. \\ \textcolor{blue}{To calculate the Hamming distance, we can} convert the integers to binary, compare the bits, and count the number of differences.\\ Here's the Python function to calculate the Hamming distance: \\ \textcolor{teal}{$\dots$ (correct code)}} \\
    \midrule
    \ding{194} & \textbf{Problem} & \makecell[l]{Write a function to get the n largest items from a dataset. \\ \texttt{def larg\_nnum(list1, n):}} \\
    \cmidrule(lr){2-3}
    & \textbf{DA} output: & \makecell[l]{Final Answer: \\ \textcolor{red}{$\dots$ (buggy code)}} \\
    \cmidrule(lr){2-3}
    & \textbf{IA} output: & \makecell[l]{The function \texttt{\`{}larg\_nnum\`{}} \textcolor{blue}{is supposed to} return the n largest items from a given list. \\ \textcolor{blue}{This can be achieved using} the \texttt{\`{}heapq.nlargest\`{}} function in Python, which returns \\ the n largest elements from the iterable specified by the key function. \\ Here's the completed function: \\ \textcolor{teal}{$\dots$ (correct code)}} \\
    \midrule
    \bottomrule
    \end{tabular}
    }
    \caption{\textbf{Cases from MBPP results}, where IA correctly answers the question while the DA baseline fails. The intentional analysis at the beginning helps with the subsequent coding by proposing feasible solutions.}
    \label{tab:case_study_mbpp}
\end{table*}

\end{document}